\documentclass{article}
\usepackage{log_2025}						% for camera-ready version

\usepackage{booktabs}						% professional-quality tables
\usepackage{multirow}						% tabular cells spanning multiple rows
\usepackage{amsfonts}						% blackboard math symbols
\usepackage{graphicx}						% figures
\usepackage{duckuments}						% sample images
\usepackage{graphicx}
\usepackage{comment}
\usepackage{amsmath}
\usepackage{amssymb}
\usepackage{soul}
\usepackage{float}
\usepackage{algorithm}
\usepackage{algorithmic}
\usepackage{subcaption}
\usepackage{booktabs}
\usepackage{tablefootnote}
\usepackage{colortbl}
\usepackage{xcolor}
\usepackage{adjustbox}
\usepackage[numbers,compress,sort]{natbib}	% for numerical citations
\title{G-Loss: Graph-Guided Fine-Tuning of Language Models}

\author[A Sharma et al.]{
Aditya Sharma$^{a}$,
Vinti Agarwal$^{a}$,
Rajesh Kumar$^{b}$,\\[0.5em]
$^{a}$BITS Pilani, India \quad
$^{b}$Bucknell University, USA\\[0.5em]
\texttt{\{p20200470, vinti.agarwal\}@pilani.bits-pilani.ac.in,} \quad
\texttt{rk042@bucknell.edu}
}

\begin{document}
\maketitle
\begin{abstract}
Traditional loss functions, including cross-entropy, contrastive, triplet, and supervised contrastive losses, used for fine-tuning pre-trained language models such as \texttt{BERT}, operate only within local neighborhoods and fail to account for the global semantic structure. We present \textit{G-Loss}, a graph-guided loss function that incorporates semi-supervised label propagation to use structural relationships within the embedding manifold. \textit{G-Loss} builds a document-similarity graph that captures global semantic relationships, thereby guiding the model to learn more discriminative and robust embeddings. We evaluate \textit{G-Loss} on five benchmark datasets covering key downstream classification tasks: \texttt{MR} (sentiment analysis), \texttt{R8} and \texttt{R52} (topic categorization), \texttt{Ohsumed} (medical document classification), and \texttt{20NG} (news categorization). In the majority of experimental setups, \textit{G-Loss} converges faster and produces semantically coherent embedding spaces, resulting in higher classification accuracy than models fine-tuned with traditional loss functions. 
\end{abstract}

\section{Introduction}

Language models, ranging from \texttt{BERT} \citep{bert} and \texttt{RoBERTa} \citep{roberta}, with millions of parameters, to large-scale models such as \texttt{GPT} \citep{GPT} and \texttt{LLaMA} \citep{llama}, with parameters in billions, have transformed natural language processing (NLP). These models follow a two-stage approach: unsupervised pre-training on large unlabeled corpora to learn general representations, followed by supervised fine-tuning with labeled data using a task-specific loss function. Fine-tuning, however, presents challenges, including the need for large labeled datasets, high computational cost, and sensitivity to task-specific objectives. Traditional fine-tuning losses, such as cross-entropy, primarily minimize prediction error while overlooking the semantic structure within the data.  

Several strategies have been explored to improve fine-tuning effectiveness. Classical approaches, such as \texttt{BERT} and \texttt{RoBERTa}, employ cross-entropy loss with a classification head attached to the encoder \citep{bert, roberta}, which enforces class separation in the embedding space. This approach, however, ignores relationships between individual samples and relies solely on label supervision \citep{scl}. Pairwise and triplet-based methods, such as \texttt{SBERT} \citep{SBERT}, address this limitation by optimizing similarity objectives that pull semantically related samples closer while pushing unrelated samples apart. These methods improve local neighborhood structure but are computationally expensive because they require generating many positive and negative pairs. More importantly, they still optimize local relationships independently, without enforcing global structural alignment. For example, ensuring that \(A\) is close to \(B\) and far from \(C\) does not guarantee a coherent global positioning of \(B\) and \(C\) unless the pair \((B, C)\) is explicitly optimized. This locality constraint often limits generalization to unseen data.  

To address these limitations, we propose a shift from local pairwise optimization to global structural alignment by modeling semantic relationships across all pairs through a graph. We introduce \textit{G-Loss}, a graph-based fine-tuning strategy that enforces both local and multi-hop consistency within a unified framework. While the approach can be applied to any encoder-based model, we focus here on transformer-based encoders for NLP tasks. \textit{G-Loss} integrates the semi-supervised Label Propagation Algorithm (LPA) \citep{LPA} into the fine-tuning process, allowing label information to diffuse from labeled to unlabeled nodes within a similarity graph. This enables the model to benefit from both explicit labels and implicit relational information, consistent with the manifold assumption in semi-supervised learning \citep{manifold_assumption}, which posits that proximity in feature space correlates with label similarity.  

A key advantage of using LPA is its parameter-free nature, which ensures scalability and efficiency across model sizes and datasets. Another distinguishing feature of our framework is its self-reinforcing nature: the graph structure and the embeddings co-evolve during fine-tuning. As embeddings improve, they reshape the graph structure, which in turn guides further embedding refinement, yielding better global alignment.  

Unlike earlier works that use LPA with static, full-dataset graphs for vision tasks \citep{iscen-lpa} or employ similarity graphs as soft targets for image classification \citep{xsample}, \textit{G-Loss} introduces three key techniques. First, it dynamically constructs mini-batch graphs using evolving embeddings, supporting inductive learning while reducing memory cost. Second, it integrates LPA directly into the loss function, eliminating the need for separate pseudo-labeling and retraining. Third, it maintains embedding consistency throughout training while streamlining the fine-tuning pipeline.  

The major contributions of this work are summarized as follows:
\begin{itemize}
    \item We propose a fine-tuning framework that integrates a graph-based loss derived from dynamically evolving graphs and the semi-supervised Label Propagation Algorithm (LPA). Unlike traditional losses that optimize only local relationships, our approach enforces global semantic consistency—the coevolution of the graph structure and the embedding space during fine-tuning yields robust, discriminative representations.

    \item We implement and evaluate the proposed framework using \texttt{BERT}, \texttt{RoBERTa}, and \texttt{DistilBERT}, demonstrating its applicability across model scales. 
The code scripts are publicly available at \footnote{
\url{https://github.com/saditya13/G-Loss-LoG}}.

    \item We conduct experiments on five benchmark datasets covering binary and multi-class classification tasks with varying class distributions. Performance is evaluated using accuracy, macro F1-score, silhouette score, and convergence time.

    \item We compare our method with standard loss functions, including supervised contrastive, cosine similarity, triplet, and cross-entropy losses.
\end{itemize}

The remainder of this paper is organized as follows. Section~\ref{related_work} reviews related literature. Section~\ref{fine_tuning_process} details the fine-tuning process with \textit{G-Loss}. Section~\ref{experimentation} describes the datasets, language models, loss baselines, and evaluation criteria. Section~\ref{results_discussion} presents and discusses the results. Section~\ref{conclusion} concludes the paper and outlines future directions.  

\section{Related work}
\label{related_work}
Loss functions play a crucial role in fine-tuning language models (LMs) by shaping how representation learning adapts to downstream tasks. Fine-tuning models such as \texttt{BERT} \citep{bert}, \texttt{RoBERTa} \citep{roberta}, and \texttt{ALBERT} \citep{albert} typically involves appending a classification head and optimizing the cross-entropy (CE) loss. Although effective for classification, CE treats training samples independently and fails to preserve structural consistency within the embedding space.  

To address this limitation, similarity-based learning methods focus on relational structure. Sentence-BERT (\texttt{SBERT}) \citep{SBERT} uses a Siamese network with triplet and cosine similarity losses for pairwise alignment of semantically related sentences. \texttt{SimCSE} \citep{simsce} introduces contrastive learning with dropout-based augmentation, promoting more discriminative and stable embeddings through positive and negative pair comparison. \texttt{SimCSE++} \citep{simcseplus} improves upon this by reducing negative-pair noise and adding a dimension-wise contrastive objective to prevent feature collapse. Despite these advances, contrastive and triplet-based losses remain locally optimized, capturing only sample-level relationships and failing to represent global semantic structure \citep{pmlr-v162-yuan22b}. \citet{ce_scl} proposed the supervised contrastive learning (SCL) objective to complement cross-entropy loss, improving class-level separation but still without explicit structural modeling.  

Graph-based approaches, in contrast, explicitly encode relational dependencies among samples. Graph Neural Networks (GNNs) extend representation learning by propagating contextual information across graph nodes \citep{gcn}. Graph Convolutional Networks (GCN) \citep{gcn} first demonstrated feature aggregation through message passing, and \texttt{TextGCN} \citep{textgcn} extended this idea to document classification via heterogeneous word-document graphs. \texttt{TensorGCN} \citep{liu2020tensorgcn} further incorporated syntactic and sequential dependencies to enrich the textual graph representations.  

Hybrid language model–graph frameworks have emerged to combine contextual embeddings with graph reasoning. For instance, \texttt{BertGCN} integrates BERT embeddings within a GCN to improve text classification \citep{bertgcn}. Other approaches include: (1) cascading models, where the language model produces embeddings subsequently processed by a GNN for classification \citep{vgcn_bert, tape, graphformer}; (2) co-training models, where LMs and GNNs are optimized jointly with shared objectives \citep{glem, leading, graphformer}; and (3) frozen LM strategies, where the language model remains fixed while only the GNN parameters are updated \citep{engine, gpeft}.  

While these methods show promise, they face two key limitations: (1) \textit{static graph construction}, where the graph structure is fixed and does not adapt to evolving embeddings, and (2) \textit{high computational overhead} from joint GNN–LM optimization. For example, \texttt{BertGCN} relies on a static, full-dataset graph and a memory bank, which significantly increase training time and memory usage.  

We propose \textit{G-Loss} to address these limitations through dynamic graph construction during fine-tuning. \textit{G-Loss} enables the language model to learn and refine semantic structure adaptively, unlike traditional fine-tuning objectives (cross-entropy, triplet, contrastive) that overlook global dependencies, or graph-based methods (\texttt{TextGCN}, \texttt{BertGCN}) that require costly static, precomputed graphs. \textit{G-Loss} bridges this gap by combining dynamic graph adaptation with efficient label propagation, allowing seamless integration of graph-structured learning into language model fine-tuning.  

\section{Proposed fine-tuning framework}
\label{fine_tuning_process}
\subsection{Task description}
The goal is to improve document-level representation learning for supervised text classification by integrating global semantic structure into the fine-tuning process of pre-trained language models.  
Formally, given a document set \(\mathcal{D} = \{d_1, d_2, \ldots, d_n\}\) and a corresponding label set \(\mathcal{Y} = \{y_1, y_2, \ldots, y_n\}\) spanning \(\mathcal{C}\) classes, the objective is to learn a mapping function
\[
f: d_i \mapsto y_i, \quad f(d_i) = \arg\max_{c \in \mathcal{C}} P(y_i = c \mid d_i; \theta),
\]
where \(\theta\) denotes the parameters of the fine-tuned language model.  

Unlike conventional fine-tuning, which optimizes sample-level prediction loss, we reformulate the task also to preserve semantic relationships among samples. This is achieved by representing each minibatch of documents as a graph \(\mathcal{G} = (\mathcal{V}_k, \mathcal{E}_k, \mathcal{X}_k)\), where nodes correspond to document embeddings, edges encode semantic similarity, and \(\mathcal{X}_k \in \mathbb{R}^{B \times d}\) contains the language model representations.  
The model is fine-tuned on the training set to minimize classification error and graph-based relational inconsistency jointly. The validation set is used to monitor convergence and for hyperparameter tuning (\(\lambda, \gamma, \sigma\)), and the held-out test set is used only for final evaluation.

%%%%%%%%%%%%%%%%%%%%%%%%%%
\subsection{G-Loss: Graph-driven loss computation}

Given the training set \(\mathcal{T} \subset \mathcal{D}\) and corresponding labels \(\mathcal{Y}\), fine-tuning proceeds over minibatches \(\{b_1, b_2, \ldots, b_m\}\).  
For each minibatch \(b_k\), documents are encoded into dense representations using a pre-trained language model \(\Phi(.)\), and a similarity-based document graph is constructed in which nodes represent documents and edge weights \(w_{ij}\) capture pairwise semantic affinity.  
To incorporate global structural consistency into the fine-tuning process, a subset of node labels is masked according to a masking ratio \(\gamma\), and the semi-supervised Label Propagation Algorithm (LPA) \citep{LPA} is applied to infer these hidden labels while keeping known labels fixed.  
The discrepancy between the inferred and true labels defines the graph-based loss \(\mathcal{L_G}\).  
The overall objective combines this graph-based loss with the standard cross-entropy loss \(\mathcal{L_{CE}}\) using a weighting factor \(\lambda\):  
\[
\mathcal{L} = \lambda \mathcal{L_G} + (1 - \lambda)\mathcal{L_{CE}}.
\]
This composite loss is optimized via backpropagation, allowing the model to jointly refine both the embeddings and the evolving graph structure during fine-tuning.

\begin{figure}[t]
    \centering    
    \includegraphics[width=1.1\linewidth]{ 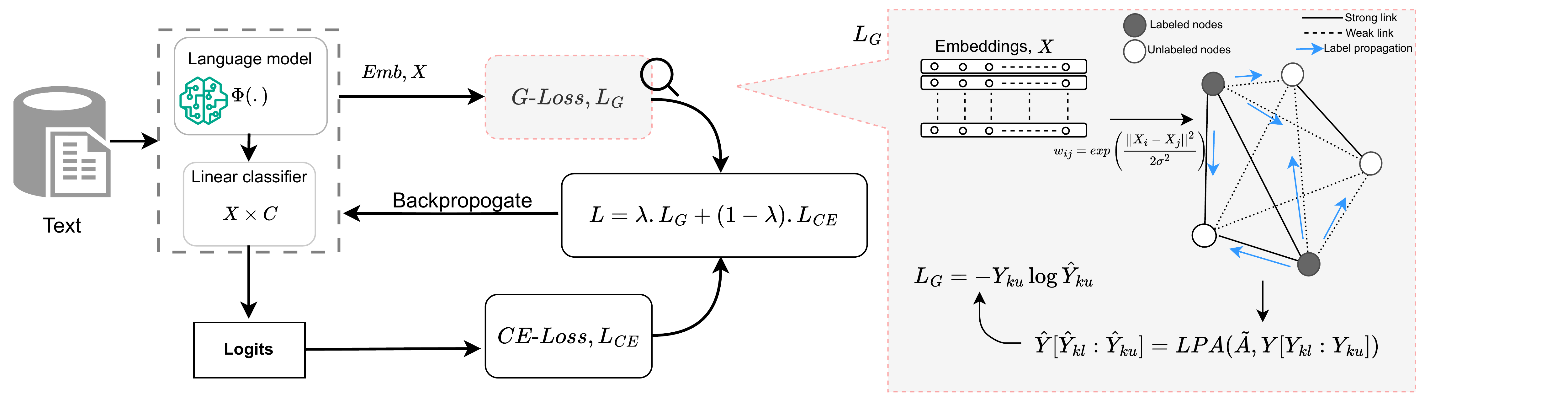}
    \caption{The language model $\Phi(.)$ encodes input text into embeddings, which are used by both a linear classifier (producing logits) and a similarity-based document graph. In the graph, nodes represent documents and edge weights $w_{ij}$ capture pairwise semantic similarity. A subset of node labels ($Y_{ku}$) is masked and inferred using the LPA, yielding predicted labels $\hat{Y}_{ku}$. The resulting graph-based loss $\mathcal{L_G}$ is combined with the cross-entropy loss $\mathcal{L_{CE}}$ using weighting factor $\lambda$. The composite loss is then backpropagated to update both the language model and classifier parameters.}
    \label{fig:gloss}  
\end{figure}

Figure~\ref{fig:gloss} illustrates this process for a single minibatch.  
The key stages of the proposed framework are described below.

\paragraph{\textbf{(Step 1) Embedding extraction}}  
Each document in minibatch \(b_k\), \(k \in \{1, 2, \ldots, m\}\), is encoded by the language model \(\Phi(.)\) into a \(d\)-dimensional embedding space:
\[
X_k = \Phi(\text{Text in } b_k),
\]
where \(X_k \in \mathbb{R}^{B \times d}\) represents contextual embeddings for the batch of size \(B\).  
These embeddings serve as node features for graph construction. The quality of these embeddings directly affects label propagation and downstream classification performance.

\paragraph{\textbf{(Step 2) Graph construction}}  
A fully connected weighted graph \(\mathcal{G}_k = (\mathcal{V}_k, \mathcal{E}_k, X_k)\) is constructed for minibatch \(b_k\).  
The edge weight between nodes \(i\) and \(j\) is computed using a Gaussian kernel:
\[
w_{ij} = \exp\!\left(-\frac{\|X_i - X_j\|^2}{2\sigma^2}\right), \quad w_{ii}=0,
\]
where \(X_i\) and \(X_j\) are document embeddings, and \(\sigma\) controls neighborhood sensitivity in kernel space.  

We evaluate two approaches for choosing \(\sigma\):  
(1) \textit{G-Loss-SQRT}, which analytically estimates \(\sigma = \sqrt{d_1 / 3}\), where \(d_1\) is the median Euclidean distance across all pairs (see Appendix~\ref{app:sigma}); and  
(2) \textit{G-Loss-O}, which optimizes \(\sigma\) through hyperparameter search.  
\textit{G-Loss-SQRT} provides a computationally efficient alternative by avoiding hyperparameter tuning overhead. The Gaussian kernel ensures that highly similar documents are strongly connected while dissimilar ones have weaker connections.  

To stabilize propagation, the adjacency matrix is symmetrically normalized:
\[
\tilde{A} = D^{-1/2} W D^{-1/2},
\]
where \(D_{ii} = \sum_{j=1}^{B} w_{ij}\) is the degree matrix. This ensures that label propagation distributes
information effectively without amplifying numerical instabilities.

\paragraph{\textbf{(Step 3) Label propagation}}  
A hyperparameter \(\gamma\) splits the minibatch node set \(\mathcal{V}_k\) into two disjoint subsets:
\begin{itemize}
    \item \(\mathcal{V}_{kl}\)(Labeled subset) : a \(\gamma\) fraction of nodes with known labels \(Y_{kl}\);
    \item \(\mathcal{V}_{ku}\) (Evaluation subset): unlabeled nodes whose labels \(Y_{ku}\) are masked.
\end{itemize}

The column-stochastic transition matrix \(T\) is defined as:
\[
T_{ij} = P(j \rightarrow i) = \frac{\tilde{A}_{ij}}{\sum_{m=1}^{B} \tilde{A}_{mj}}.
\]
Using the closed-form solution of LPA \citep{LPA}, the inferred labels for the evaluation subset are:
\[
\hat{Y}_{ku} = (I - T_{uu})^{-1} T_{ul} Y_{kl},
\]
where \(I\) is the identity matrix, \(T_{uu}\) is the submatrix for unlabeled nodes, and \(T_{ul}\) captures transitions from labeled to unlabeled nodes.

\paragraph{\textbf{(Step 4) Loss computation}}  
The graph-based loss \(\mathcal{L_G}\) measures the discrepancy between the inferred labels \(\hat{\mathbf{Y}}_{ku}\) and the true labels \(\mathbf{Y}_{ku}\) over the evaluation subset of size \(B_e = (1-\gamma)B\):
\begin{equation}
\mathcal{L_G} = -\frac{1}{B_e} \sum_{j=1}^{B_e} \sum_{c=1}^{C} 
y_{ku, j}^c \log \big( \hat{y}_{ku, j}^c \big),
\label{eq:loss_main}
\end{equation}
where \(y_{ku,j}^c\) and \(\hat{y}_{ku,j}^c\) denote the ground-truth and predicted probabilities for class \(c\) of the \(j^{th}\) evaluation node, and \(C\) is the number of classes.

\paragraph{\textbf{(Step 5) Loss integration}}  
The final objective unifies the previously defined components by combining the graph-based loss \(\mathcal{L_G}\) and the cross-entropy loss \(\mathcal{L_{CE}}\):
\begin{equation}
\mathcal{L} = \lambda\, \mathcal{L_G} + (1 - \lambda)\, \mathcal{L_{CE}},
\end{equation}
where \(\lambda \in [0,1]\) balances the contribution of both terms.  
The cross-entropy component is expressed as:
\begin{equation}
\mathcal{L_{CE}} = -\frac{1}{B} \sum_{i=1}^{B} \sum_{c=1}^{C} y_{i,c}\log \hat{y}_{i,c}.
\end{equation}
Here, ${B}$ denotes the minibatch size.
\paragraph{\textbf{(Step 6) Model optimization and dynamic graph updating}}  
The language model and classifier parameters are jointly optimized via backpropagation using the composite loss \(\mathcal{L}\):
\begin{equation}
\mathbf{X}_k' \leftarrow \mathbf{X}_k - \eta \nabla \mathcal{L},
\end{equation}
where \(\eta\) denotes the learning rate.  
As embeddings evolve during fine-tuning, the graph structure is recomputed after each update:
\begin{equation}
\mathbf{W}_{ij}' = \exp\!\left(-\frac{\|\mathbf{X}_i' - \mathbf{X}_j'\|^2}{2\sigma^2}\right).
\end{equation}
This dynamic adaptation ensures that the similarity graph remains consistent with the evolving embedding space, enabling more coherent representations and improved classification performance.  

This completes one iteration of fine-tuning under the \textit{G-Loss} framework. A detailed algorithm of the approach is provided in section \ref{app:algorithm} of the appendix.  
The next section describes the experimental setup, benchmark datasets, and evaluation protocol used to assess its effectiveness.

%%%%%%%%%%%%%%%%%%%%%%%%%%%%%%%%%%%%%%%%%%%
%%%%PROOFREAD TILL HERE %%%%%%%
%%%%%%%%%%%%%%%%%%%%%%%%%%%%%%%%%%%%%%%%%%%
\section{Experimental setup}
\label{experimentation}

\subsection{Datasets and downstream tasks}
We evaluate G-Loss on five widely used English text classification benchmarks.  
\ul{\textit{Movie Review (MR)}} \citep{Pang+Lee:2004}: binary sentiment classification (positive/negative).  
\ul{\textit{R8}} and \ul{\textit{R52}} \citep{Lewis:1997}: topic classification derived from Reuters-21578 with 8 and 52 categories, respectively.  
\ul{\textit{Ohsumed}} \citep{Hersh:1994}: medical text classification using MEDLINE abstracts labeled under 23 MeSH categories.  
\ul{\textit{20 Newsgroups (20NG)}} \citep{Lang:1995}: news articles grouped into 20 discussion topics.  
For consistency, we adopt the data splits released by \citet{bertgcncode}, ensuring identical data partitions. Dataset statistics are summarized in Table~\ref{table:datasets}.

\begin{table}[ht]
    \centering 
    \caption{Dataset statistics used for evaluation. The Train, Validation, and Test columns denote the number of documents in each split.}
    \begin{tabular}{l r r r c r}
    \hline
    \textbf{Dataset} & \textbf{\# Docs} & \textbf{Train} & \textbf{Validation} & \textbf{Test} & \textbf{\# Classes} \\ \hline
    MR & 10662 & 6397 & 711 & 3554 & 2 \\ 
    R8 & 7674 & 4936 & 549 & 2189 & 8 \\ 
    R52 & 9100 & 5865 & 667 & 2568 & 52 \\ 
    20NG & 18846 & 10182 & 1132 & 7532 & 20 \\ 
    Ohsumed & 7400 & 3021 & 336 & 4043 & 23 \\ \hline
    \end{tabular} 
    \label{table:datasets}
\end{table}

\subsection{Language models and traditional losses}
\label{language_models}

\paragraph{\ul{\textbf{Language Models}}}
\textit{G-Loss} is compatible with any encoder-based transformer.  
We evaluate it using three models of varying size and complexity:  
\texttt{BERT-base-uncased} (12 layers, 768 hidden units, 110M parameters) \citep{bert},  
\texttt{RoBERTa-large} (24 layers, 1024 hidden units, 356M parameters) \citep{roberta}, and  
\texttt{DistilBERT-base-uncased} (6 layers, 768 hidden units, 66M parameters) \citep{distilbert}.  

\paragraph{\ul{\textbf{Traditional loss functions}}}
\label{baselines}We assess \textit{G-Loss} under two configurations: \ul{integrated} and \ul{standalone}.  

In the \textbf{integrated setup}, the hybrid loss jointly optimizes the language model and classifier parameters, following the framework in Figure~\ref{fig:gloss}.  
This setting aligns with standard fine-tuning approaches such as BERT, RoBERTa, and BertGCN \citep{bertgcn, ce_scl}.  

In the \textbf{standalone setup}, only the language model parameters are updated using the computed loss, while the classifier is trained separately on the resulting embeddings.  

\begin{itemize}
    \item \textbf{Integrated:} \textit{G-Loss} (with CE) is compared against cross-entropy alone, and the hybrid Supervised Contrastive + CE loss (SCL+CE) \citep{ce_scl}.
    \item \textbf{Standalone:} \textit{G-Loss} is compared against triplet loss \citep{triplet}, Supervised Contrastive Loss (SCL) \citep{scl}, and cosine-similarity loss.
\end{itemize}
Mathematical formulations for all baselines are provided in Appendix~\ref{mathematical_formula}.

\subsection{Fine-tuning convergence and hyperparameter tuning}

\paragraph{\ul{\textbf{Evaluation strategies and early stopping}}}
\label{early-stopping}
In the \ul{integrated} configuration, training is monitored using the \ul{macro F1-score} on the validation set.  
Early stopping is applied if no improvement occurs within a fixed patience window.  
The classifier is a linear layer of size \(E \times C\), where \(E\) is the embedding dimension and \(C\) the number of classes.

In the \ul{standalone} configuration, convergence is monitored using the \ul{macro-silhouette score} \citep{macrosil} computed on validation labels (formula in Appendix~\ref{app:silhouette}).  
Unlike the standard silhouette score, this variant accounts for class imbalance.  
After fine-tuning, a linear classifier is trained on the frozen embeddings, and test predictions are obtained from this downstream model.  
This setup avoids training a classification head jointly with the language model, reducing parameter count and computation.

\paragraph{\ul{\textbf{Hyperparameter selection and final classifier}}}
Hyperparameter tuning is performed using \texttt{Optuna}.  
For \textit{G-Loss}, we tune: Gaussian kernel width \(\sigma \in [0.1, 10]\), learning rate \(\eta \in \{1\mathrm{e}{-5}, 2\mathrm{e}{-5}, 3\mathrm{e}{-5}, 4\mathrm{e}{-5}, 5\mathrm{e}{-5}\}\), label masking ratio \(\gamma \in [0.1, 0.9]\), and weighting factor \(\lambda \in [0.1, 0.9]\).  
Baseline losses use comparable optimization: temperature \(\tau \in [0.01, 1]\) and learning rate \(\eta\) for SCL, and learning rate \(\eta\) for cross-entropy, cosine-similarity, and triplet losses.  
\textcolor{black}{Details on the hyperparameter search are provided in Appendix~\ref{app:hyperparam}.  
All experiments were conducted on a single NVIDIA A100 80GB GPU.}

\paragraph{\ul{\textbf{Evaluation metrics and protocol}}}
We report test accuracy and macro F1-score as primary metrics, ensuring comparability with prior work such as TextGCN, TensorGCN, and BertGCN.  
Macro F1 is particularly relevant for imbalanced datasets (e.g., R52 and Ohsumed).  
\textcolor{black}{To assess computational efficiency, we also record fine-tuning time to convergence and per-epoch training time for \textit{G-Loss} and all baseline losses, as shown in Table~\ref{standalone}.}  

%%%%%%%%%%%%%%%%%%%%%%%%%%%%%%%%%%%%%%%%%%%
% PROOF READ TILL HERE!!!! 
%%%%%%%%%%%%%%%%%%%%%%%%%%%%%%%%%%%%%%%%%%%
\section{Results and discussion}
\label{results_discussion}
\subsection{\textit{G-Loss} with integrated pipeline}
We experimentally compare the proposed \textit{G-Loss} with strong baselines: CE loss and SCL+CE loss (section~\ref{baselines}). For robust assessment, we used three distinct language model architectures (section~\ref{language_models}).

\colorlet{SoftYellow}{Yellow!40} 
\begin{table*}[ht]
\caption{\textcolor{black}{Accuracy(\%)/macro F1-score (\%) obtained while fine-tuning different language model variants across multiple datasets. The values highlighted in \colorbox{lightgray}{gray} and \colorbox{SoftYellow}{yellow} denote the best and second-best performance, respectively. We report the mean and variance of three different seeds.}}
\centering
\scalebox{0.53}{
\begin{tabular}{@{}llccccc@{}}
\toprule
\textbf{Model} & \textbf{Loss} & \textbf{MR} & \textbf{R8} & \textbf{R52} & \textbf{20NG} & \textbf{Ohsumed} \\ \midrule
\multirow{4}{*}{\begin{tabular}[c]{@{}l@{}}BERT-base \\ -uncased\end{tabular}} 
& CE & 85.64$\pm$0.81 / 85.64$\pm$0.81 & 97.57$\pm$0.16 / 93.90$\pm$0.84 & 96.29$\pm$0.15 / 84.42$\pm$0.60 & 83.98$\pm$0.12 / 83.49$\pm$0.18 & 71.09$\pm$0.31 / 63.40$\pm$0.83 \\
& SCL + CE\cite{ce_scl} & 85.97$\pm$0.50 / 85.95$\pm$0.50 & \colorbox{SoftYellow}{97.88$\pm$0.17 / 93.96$\pm$0.57} & \colorbox{SoftYellow}{96.18$\pm$0.14 / 83.75$\pm$0.77} & \colorbox{SoftYellow}{84.39$\pm$0.10 / 83.99$\pm$0.13} & \colorbox{lightgray}{71.28$\pm$0.17 / 63.64$\pm$0.20} \\
& GLoss-SQRT + CE & \colorbox{SoftYellow}{86.17$\pm$0.35 / 86.17$\pm$0.35} & 97.67$\pm$0.19 / 94.34$\pm$0.66 & 96.06$\pm$0.18 / 83.78$\pm$0.48 & 84.01$\pm$0.56 / 83.52$\pm$0.62 & 70.55$\pm$0.18 / 62.49$\pm$0.73 \\
& GLoss-O + CE & \colorbox{lightgray}{86.63$\pm$0.12 / 86.61$\pm$0.12} & \colorbox{lightgray}{97.91$\pm$0.20 / 94.53$\pm$0.76} & \colorbox{lightgray}{96.37$\pm$0.40 / 84.55$\pm$0.38} & \colorbox{lightgray}{84.71$\pm$0.24 / 84.17$\pm$0.26} & \colorbox{SoftYellow}{71.25$\pm$0.59 / 63.80$\pm$0.82} \\

  \hline
  
\multirow{4}{*}{\begin{tabular}[c]{@{}l@{}}RoBERTa \\ -large\end{tabular}} 
& CE & 89.92$\pm$1.07 / 89.92$\pm$1.07 & \colorbox{lightgray}{97.75$\pm$0.20 / 93.46$\pm$0.37} & 95.91$\pm$0.56 / 83.71$\pm$1.09 & 84.89$\pm$0.36 / 84.28$\pm$0.39 & \colorbox{SoftYellow}{75.03$\pm$0.24 / 67.95$\pm$0.59} \\
& SCL + CE \cite{ce_scl} & {90.21$\pm$0.91/ 90.20$\pm$0.92} & 97.24$\pm$0.19 / 93.11$\pm$0.60 & 95.95$\pm$0.45 / 82.33$\pm$1.37 & \colorbox{lightgray}{85.31$\pm$0.26 / 84.90$\pm$0.25} & {73.67$\pm$0.33} / {65.55$\pm$0.42 } \\ 
& G-Loss-SQRT + CE & \colorbox{SoftYellow}{90.53$\pm$0.51 / 90.53$\pm$0.51} & 97.26$\pm$0.12 / 93.02$\pm$0.21 & \colorbox{SoftYellow}{96.14$\pm$0.47 / 84.31$\pm$1.02} & 85.06$\pm$0.28 / 84.61$\pm$0.58 & 74.82$\pm$0.52 / 66.55$\pm$0.74 \\
& G-Loss-O + CE & \colorbox{lightgray}{90.87$\pm$0.62 / 90.87$\pm$0.62} & \colorbox{SoftYellow} {97.49$\pm$0.05 / 93.04$\pm$0.12} & \colorbox{lightgray}{96.44$\pm$0.63 / 84.34$\pm$0.85} & \colorbox{SoftYellow}{85.19$\pm$0.31 / 84.23$\pm$0.73} & \colorbox{lightgray}{75.35$\pm$0.42 / 68.58$\pm$1.02} \\ 

\hline

\multirow{4}{*}{\begin{tabular}[c]{@{}l@{}}DistilBERT \\ -base\end{tabular}} 
& CE & 84.85$\pm$0.23 / 84.85$\pm$0.23 & 97.48$\pm$0.11 / 93.02$\pm$0.48 & 95.78$\pm$0.04 / 83.64$\pm$0.69 & 83.15$\pm$0.61 / 82.69$\pm$0.56 & 69.50$\pm$0.34 / 61.06$\pm$0.29 \\
& SCL + CE \cite{ce_scl} & 84.16$\pm$0.46 / 84.16$\pm$0.46 & \colorbox{SoftYellow}{97.65$\pm$0.25 / 93.79$\pm$0.43} & \colorbox{lightgray}{96.17$\pm$0.10} / 83.82$\pm$1.62 & \colorbox{SoftYellow}{83.49$\pm$0.20 / 83.18$\pm$0.12} & \colorbox{SoftYellow}{70.02$\pm$0.35} / 60.56$\pm$0.80 \\
& GLoss-SQRT + CE & \colorbox{lightgray}{85.14$\pm$0.33 / 85.14$\pm$0.34} & 97.46$\pm$0.17 / 93.62$\pm$0.69 & \colorbox{SoftYellow}{96.13$\pm$0.34 / 83.93$\pm$0.35} & 83.61$\pm$0.23 / 83.16$\pm$0.20 & 69.93$\pm$0.38 / \colorbox{SoftYellow}{61.39$\pm$0.21} \\
& GLoss-O + CE & \colorbox{SoftYellow}{85.10$\pm$0.45 / 85.10$\pm$0.45} & \colorbox{lightgray}{97.71$\pm$0.12 / 93.61$\pm$0.22} & 96.07$\pm$0.04 / \colorbox{lightgray}{84.44$\pm$0.43} & \colorbox{lightgray}{83.64$\pm$0.28 / 83.23$\pm$0.27} & \colorbox{lightgray}{70.16$\pm$0.21 / 62.41$\pm$0.34} \\

\hline
\end{tabular}
}
\label{main_results}
\end{table*}

Table~\ref{main_results} shows classification accuracy and macro F1-score across all datasets and language model variants.  
\textit{G-Loss} variants closely match or marginally surpass other baselines regardless of the underlying language model, achieving improvements of $0.03\%-1.02\%$ in accuracy and $0.16\%-1.35\%$ in macro F1-score compared to the second-best-performing loss, across all datasets. 
Notably, \textit{G-Loss} variants generalize across different language model families, indicating its effectiveness in enhancing a variety of language models. 
While SCL+CE remains competitive on some datasets, it underperforms on large models such as RoBERTa-large. These findings confirm that our graph-based loss function consistently improves language model fine-tuning for text classification.

\subsection{\textit{G-Loss} as a standalone Objective}
Table \ref{standalone} presents the standalone evaluation of \textit{G-Loss} compared with traditional loss functions (detailed in section \ref{baselines}). We report results for \textit{G-Loss-O}, the best-performing variant. Overall, \textit{G-Loss-O} delivers performance that is competitive or superior to, traditional objectives. In particular, it achieves the highest Macro F1 on challenging datasets, such as Ohsumed ($61.43\%$), and attains the highest silhouette scores across most benchmarks. These results demonstrates that beyond improving predictive accuracy, \textit{G-Loss-O} enforces the formation of more semantically coherent embedding spaces.
% \textcolor{black}{Notably, \textit{G-Loss-O} converges in fewer epochs (e.g., $23$ on R8) compared to alternative loss functions, highlighting its superior training efficiency. The average per-epoch training time and total time to convergence show that \textit{G-Loss} maintains comparable per-epoch computational cost to existing baselines while achieving faster overall convergence.} The table \ref{tab:silhouette_time} summarizes the quantitative comparison of \textit{G-Loss-O} and SCL (the closest competitor) across five benchmark datasets in terms of both representation quality and training efficiency. In addition, Appendix~\ref{t-test} reports a paired t-test evaluation between \textit{G-Loss} and all baseline losses. The results show consistent statistical significance in favor of \textit{G-Loss}, with every comparison yielding $p < 0.05$.

Building on these performance gains, \textit{G-Loss} also demonstrates notable training efficiency. It converges faster, requiring fewer epochs to reach optimal performance (e.g., $23$ epochs on R8). Importantly, despite incorporating graph-based components, the average per-epoch training time remains comparable to existing baselines, leading to faster overall convergence and improved training time. 
We further analyze the computational efficiency of \textit{G-Loss}. Figure~\ref{fig:epoch_breakdown} in the appendix provides a detailed timing breakdown per epoch for \textit{G-Loss} and competing baselines. \textit{G-Loss} requires $9.2$ seconds per epoch (BERT forward pass: $8.95$ sec; graph construction: $0.18$ sec; LPA operations: $0.07$ sec), compared to Supervised Contrastive Loss ($10.41$ sec), Triplet Loss ($9.15$ sec), and Cosine Similarity Loss ($9.09$ sec). This breakdown confirms that additional graph-based components - graph construction, normalization, and label propagation - introduce negligible computational overhead during training. 

Adding to this, table~\ref{tab:silhouette_time} provides a quantitative comparison between \textit{G-Loss-O} and SCL (the closest competitor) across five benchmark datasets, evaluating both representation quality and training efficiency. Complementing this analysis, Appendix~\ref{app:significance} reports paired t-test results comparing \textit{G-Loss} against all baseline losses. These statistical tests confirm the robustness of the improvements delivered by \textit{G-Loss}, with all comparisons yielding $p<0.05$.

\subsection{GLUE benchmark performance}
\begin{figure*}[t]
\centering
\begin{tabular}{@{}p{.45\textwidth} p{.5\textwidth}@{}}
% ---------------- LEFT: TABLE ----------------
\begin{minipage}[t]{\linewidth}\vspace{0pt}\scriptsize
\captionof{table}{Accuracy (\%) across large-scale GLUE benchmarks SST-2 and QNLI, highlighting the strong generalization ability of \textit{G-Loss}.}
\setlength{\tabcolsep}{3pt}
\centering
\begin{tabular}{llcc}
\toprule
Model & Loss & SST2 & QNLI \\
\midrule
\multirow{3}{*}{\begin{tabular}[c]{@{}l@{}}BERT-base \\ -uncased\end{tabular}} 
    & CE         & 91.20 & 89.08 \\
    & SCL+CE     & \colorbox{SoftYellow}{92.10} & \colorbox{LightGray}{90.06} \\
    & GLoss-O+CE & \colorbox{LightGray}{93.32} & \colorbox{SoftYellow}{89.62} \\ \midrule

\multirow{3}{*}{\begin{tabular}[c]{@{}l@{}}RoBERTa \\ -large\end{tabular}} 
    & CE         & 94.84 & 93.71 \\
    & SCL+CE     & \colorbox{SoftYellow}{95.83} & \colorbox{SoftYellow}{94.01} \\
    & GLoss-O+CE & \colorbox{LightGray}{95.88} & \colorbox{LightGray}{94.27} \\ \midrule

\multirow{3}{*}{\begin{tabular}[c]{@{}l@{}}DistilBERT \\ -base\end{tabular}} 
    & CE         & 90.14 & \colorbox{SoftYellow}{87.86} \\
    & SCL+CE     & \colorbox{SoftYellow}{90.36} & 87.18 \\
    & GLoss-O+CE & \colorbox{LightGray}{91.06} & \colorbox{LightGray}{88.34} \\
\bottomrule
\end{tabular}
\label{tab:glue}
\end{minipage}
&
% ---------------- RIGHT: FIGURE ----------------
\begin{minipage}[t]{\linewidth}\vspace{0pt}\scriptsize\centering
\includegraphics[width=\linewidth]{ 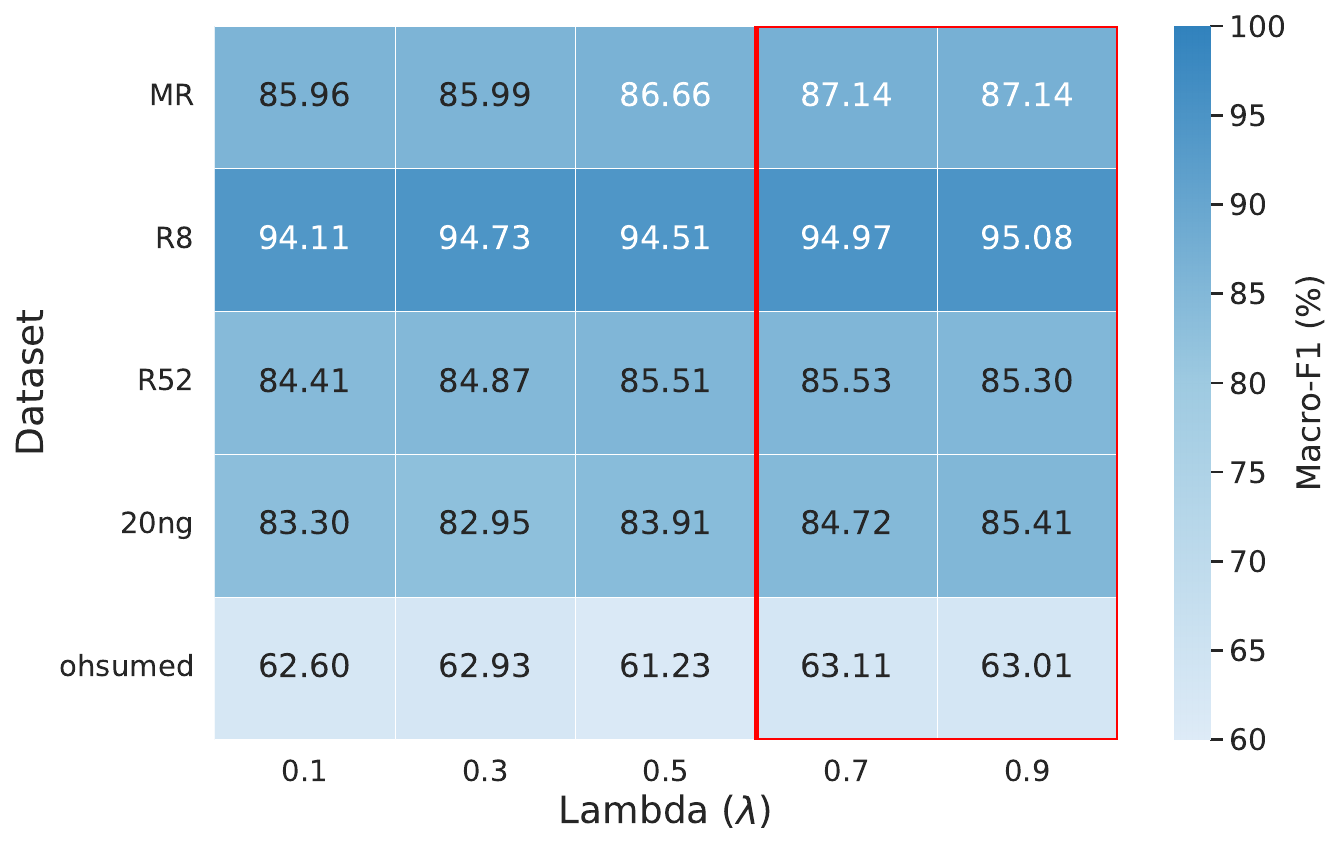}
    \caption{Loss weighting factor Lambda $(\lambda)$ vs performance of \textit{G-Loss} on \texttt{BERT-base-uncased}}
    \label{fig:lambda}
\end{minipage}

\end{tabular}

\end{figure*}

To assess the scalability and generalization of \textit{G-Loss} on large-scale natural language understanding tasks, we conduct experiments on two GLUE benchmark datasets: SST-2 and QNLI (Table~\ref{tab:glue}). SST-2 is a binary sentiment classification task from the Stanford Sentiment Treebank with approximately $67k$ training examples. QNLI, derived from the Stanford Question Answering Dataset, is a large-scale natural language inference task with approximately $105k $ sentence pairs in training set and requires sentence-level semantic reasoning. Across both tasks and different language models, \textit{G-Loss-O} outperforms both cross-entropy (CE) and supervised contrastive loss (SCL). For example, on SST-2 with \texttt{BERT-base-uncased}, \textit{G-Loss-O+CE} improves accuracy from $92.10\%$ to $93.32\%$ ($+1.22\%$). Similar improvements are observed for RoBERTa and DistilBERT, demonstrating that \textit{G-Loss} scales effectively to larger datasets and maintains strong performance across diverse architectures. 
% \textcolor{red}{add details of GLUE split experiment setup}

\begin{table}
\centering
\caption{\textcolor{black}{Comparison of \textit{G-Loss-O} with traditional losses using \texttt{BERT-base-uncased}. \textit{G-Loss-O} achieves superior or competitive accuracy and macro-F1, produces well-separated embeddings (higher silhouette scores), and converges in fewer epochs, demonstrating both effectiveness and efficiency. The highlighted colors are the same as in Table \ref{main_results}}}
\scalebox{0.77}{
\begin{tabular}{clcccccc}
\hline
\textbf{Dataset} & \textbf{Loss} & \textbf{Accuracy (\%)} & \textbf{Macro F1 (\%)} & \textbf{\begin{tabular}[c]{@{}c@{}}Test macro\\ silhouette score\end{tabular}} &  \textbf{\begin{tabular}[c]{@{}c@{}}Total train\\ time \\(sec)\end{tabular}} & \textbf{\begin{tabular}[c]{@{}c@{}}Avg. time\\ per epoch \\ (sec)\end{tabular}} & \textbf{\begin{tabular}[c]{@{}c@{}}Early stopping\\ epoch\end{tabular}} \\ \hline
\multirow{4}{*}{\textbf{MR}} & SCL & 86.18$\pm$0.49 & 86.18$\pm$0.49 & \colorbox{lightgray}{0.5725} & 1258.41 & 33.04 & 35 \\
 & Triplet & \colorbox{lightgray}{86.30$\pm$0.24} & \colorbox{lightgray}{86.29$\pm$0.24} & 0.5331 &941.55 &\colorbox{lightgray}{27.50} & 29 \\
 & Cos-sim & 86.08$\pm$0.45 & 86.08$\pm$0.45 & 0.4479 & \colorbox{lightgray}{654.56} & \colorbox{SoftYellow}{27.70} & \colorbox{lightgray}{20} \\
 & GLoss-O & \colorbox{SoftYellow}{86.22$\pm$0.25} & \colorbox{SoftYellow}{86.22$\pm$0.25} & 0.5507 & \colorbox{SoftYellow}{810.85} & 27.77 & \colorbox{SoftYellow}{25} \\ \hline
 
\multirow{4}{*}{\textbf{R8}} & SCL & \colorbox{SoftYellow}{97.80$\pm$0.07} & 93.54$\pm$0.52 & 0.4869 & 1088.58&25.56 & {38} \\
 & Triplet & 97.30$\pm$0.05 & 93.35$\pm$0.98 & 0.6581 & \colorbox{SoftYellow}{727.85}&\colorbox{lightgray}{21.19} & \colorbox{SoftYellow}{30} \\
 & Cos-sim & {97.74$\pm$0.12} & \colorbox{SoftYellow}{93.97$\pm$0.11} & 0.7383 &4960.64 & \colorbox{SoftYellow}{21.65}& 200 \\
 & GLoss-O & \colorbox{lightgray}{97.83$\pm$0.13} & \colorbox{lightgray}{93.95$\pm$0.31} & \colorbox{lightgray}{0.7870} &\colorbox{lightgray}{572.29} & 21.77 & \colorbox{lightgray}{23} \\ \hline
 
\multirow{4}{*}{\textbf{R52}} & SCL & 95.11$\pm$0.74 & \colorbox{SoftYellow}{79.72$\pm$1.02} & 0.4561 & \colorbox{SoftYellow}{1170.50} &30.41 & \colorbox{lightgray}{34} \\
 & Triplet & {95.08$\pm$0.60} & {77.78$\pm$0.63} & 0.4662 & 5496.14&\colorbox{lightgray}{25.05} & 189 \\
 & Cos-sim & \colorbox{SoftYellow}{95.21$\pm$1.11} & 79.57$\pm$2.01 & 0.3972 &1925.82 &\colorbox{SoftYellow}{25.79} & 64 \\
 & GLoss-O & \colorbox{lightgray}{95.79$\pm$0.74} & \colorbox{lightgray}{79.92$\pm$1.71} & \colorbox{lightgray}{0.4824} & \colorbox{lightgray}{1095.93}&25.87 & \colorbox{SoftYellow}{39} \\ \hline
 
\multirow{4}{*}{\textbf{Ohsumed}} & SCL & 68.54$\pm$0.69 & 57.97$\pm$0.93 &  0.1373 & \colorbox{SoftYellow}{691.03}&15.70 &  \colorbox{lightgray}{39} \\
 & Triplet & 68.68$\pm$0.40 & 58.34$\pm$0.55 & 0.1222 &1576.44 &\colorbox{lightgray}{13.09} & 105 \\
 & Cos-sim & \colorbox{SoftYellow}{69.33$\pm$0.38} & \colorbox{SoftYellow}{59.58$\pm$0.40} & 0.1393 & 3064.31&\colorbox{SoftYellow}{13.42} & 200 \\
 & GLoss-O & \colorbox{lightgray}{70.25$\pm$0.37} & \colorbox{lightgray}{61.43$\pm$1.16} & \colorbox{lightgray}{0.1468} & \colorbox{lightgray}{633.50}&13.50 & \colorbox{SoftYellow}{41} \\ \hline
 
\multirow{4}{*}{\textbf{20NG}} & SCL & \colorbox{SoftYellow}{84.45$\pm$0.79} & \colorbox{SoftYellow}{83.66$\pm$0.57} & 0.5507 &11671.10 & 52.76& 200\\
 & Triplet & 84.23$\pm$0.30 & 83.80$\pm$0.28 & 0.4538 &\colorbox{lightgray}{8516.99} & \colorbox{lightgray}{43.46}& \colorbox{lightgray}{171} \\
 & Cos-sim &  84.11$\pm$0.30 & 83.67$\pm$0.31 & 0.5524 &10155.44 &\colorbox{SoftYellow}{45.10} & 200  \\
 & GLoss-O & \colorbox{lightgray}{84.94$\pm$0.16} & \colorbox{lightgray}{84.26$\pm$0.22} & \colorbox{lightgray}{0.5849} &\colorbox{SoftYellow}{8596.50} &46.10 & \colorbox{SoftYellow}{180} \\ \hline
\end{tabular}}
\label{standalone}
\end{table}

\begin{table*}
\centering
\caption{
Comparison of \textit{G-Loss-O} and SCL (the closest competitor) across silhouette scores and training efficiency. Higher silhouette scores indicate better cluster separability; lower time per epoch indicates faster optimization.
}
\scalebox{0.98}{
\begin{tabular}{@{}lccccc@{}}
\toprule
\textbf{Dataset} & \multicolumn{2}{c}{\textbf{Silhouette Score (↑)}} & 
\multicolumn{2}{c}{\textbf{Average Time per Epoch (s, ↓)}} & 
\textbf{Observations} \\ 
\cmidrule(lr){2-3} \cmidrule(lr){4-5}
 & \textbf{G-Loss-O} & \textbf{SCL} & \textbf{G-Loss-O} & \textbf{SCL} &  \\ 
\midrule
\textbf{MR}      & 0.5507 & 0.5725 & 27.77 & 33.04 & $-3.8\%$ Sil $\downarrow$ / \textbf{16.0\% faster} \\
\textbf{R8}      & 0.7870 & 0.4869 & 21.77 & 25.56 & \textbf{+61.6\% Sil $\uparrow$} / 14.8\% faster \\
\textbf{R52}     & 0.4824 & 0.4561 & 25.87 & 30.41 & \textbf{+5.8\% Sil $\uparrow$} / 14.9\% faster \\
\textbf{Ohsumed} & 0.1468 & 0.1373 & 13.50 & 15.70 & \textbf{+6.9\% Sil $\uparrow$} / 14.0\% faster \\
\textbf{20NG}    & 0.5849 & 0.5507 & 46.10 & 52.76 & \textbf{+6.2\% Sil $\uparrow$} / 12.6\% faster \\
\bottomrule
\end{tabular}}
\label{tab:silhouette_time}
\end{table*}

\subsection{Comparison with SOTA models}
\label{sota-comparison}
Table \ref{sota} provides a comparison of our \textit{G-Loss} fine-tuned language models against several state-of-the-art baselines.
The baselines include three major paradigms in text classification: (i) graph-based methods (TextGCN \cite{textgcn}, TensorGCN \cite{liu2020tensorgcn}), (ii) transformer-based language models (BERT-base-uncased, RoBERTa-large language model, reported by \citet{bertgcn}), and (iii) hybrid models that integrate language models with graph classification (Bert-GCN and RoBERTa-GCN versions of BertGCN \cite{bertgcn}). All baseline results are taken from their respective published scores on the benchmark datasets used in this study. To ensure fairness across baselines, accuracy is reported as the primary evaluation metric.

\textit{G-Loss} delivers substantial gains over both standard pre-trained models and graph-based baselines. When combined with BERT-base, \textit{G-Loss} consistently outperforms vanilla BERT across all datasets, achieving up to ($+1.84\%$) improvement on MR and ($+0.98\%$) on Ohsumed, highlighting the benefits of graph-driven structural supervision. Notably, \textit{G-Loss} paired with RoBERTa-large attains new state-of-the-art results on MR ($90.82$), R52 ($96.65$), and Ohsumed ($75.76$), while remaining competitive on the remaining two datasets. In contrast to BertGCN and RoBERTa-GCN, which require full-graph construction and incur significant memory costs, \textit{G-Loss} achieves similar or superior accuracy with a lightweight, mini-batch dynamic graph, ensuring scalability and inductive generalization. Additionally, BertGCN's simultaneous co-training of BERT and GCN substantially increases computational resource requirements. 
Above results demonstrate that \textit{G-Loss} not only strengthens language model fine-tuning but also bridges the gap between graph-based and transformer-based paradigms more efficiently.

\begin{table}[t]
\centering
\caption{Performance comparison (accuacy in \%) with SOTA across benchmark datasets. \textit{G-Loss-O} consistently improves over graph-based methods and BERT, achieving competitive results with BertGCN while avoiding full-graph overhead through dynamic mini-batch graph construction.}
\begin{tabular}{@{}lccccc@{}}
\toprule
\textbf{Model} & \textbf{MR} & \textbf{R8} & \textbf{R52} & \textbf{Ohsumed} & \textbf{20NG} \\ \midrule
TextGCN & 76.74 & 97.07 & 93.56 & 68.36 & 86.34 \\
TensorGCN & 77.91 & 98.04 & 95.05 & 70.11 & \colorbox{SoftYellow}{87.74}  \\
BERT-base & 85.30 & 97.80 & 96.40 & 70.50 & 85.70 \\
RoBERTa-large & 89.40 & 97.80 & 96.20 & 70.70 & 83.80 \\
Bert-GCN & 86.00 & {98.10} & \colorbox{SoftYellow}{96.60} & \colorbox{SoftYellow}{72.80} & 89.30 \\
RoBerta-GCN & \colorbox{SoftYellow}{89.70 }& \colorbox{lightgray}{98.20} & {96.10} & \colorbox{SoftYellow}{72.80} & \colorbox{lightgray}{89.50} \\
G-Loss + BERT-base & {87.14} & {98.04} & {96.48} & {71.48} & 85.13 \\ 
G-Loss + RoBERTa-large & \colorbox{lightgray}{90.82} & \colorbox{SoftYellow}{98.18} & \colorbox{lightgray}{96.65} & \colorbox{lightgray}{75.76} & 85.33 \\ \bottomrule
\end{tabular}
\label{sota}
\end{table}

\section{Ablation study}
We conduct an ablation study to explore the effectiveness of our proposed graph-based loss function, utilizing the \texttt{Bert-base-uncased} language model due to its lightweight architecture and reduced computational overhead. 

\begin{figure}[]
   \centering
   \begin{subfigure}[b]{0.48\textwidth}
       \centering
       \includegraphics[width=\textwidth, height=1.5in]{ 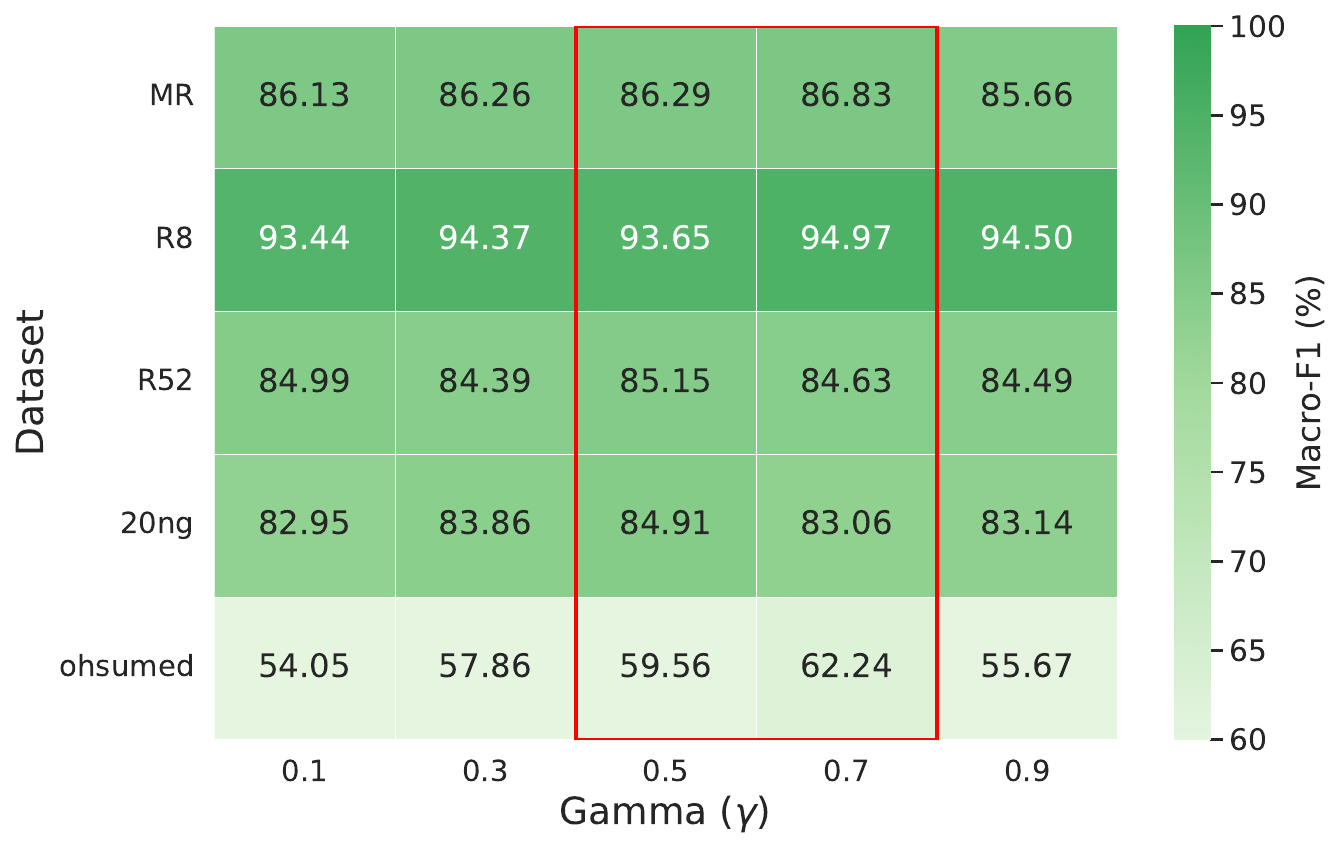}
       \caption{}
       % Label hiding ratio $\gamma$ vs Macro-F1
       \label{fig:gamma}
   \end{subfigure}
   \hfill
   \begin{subfigure}[b]{0.48\textwidth}
       \centering
       \includegraphics[width=\textwidth, height=1.5in]{ 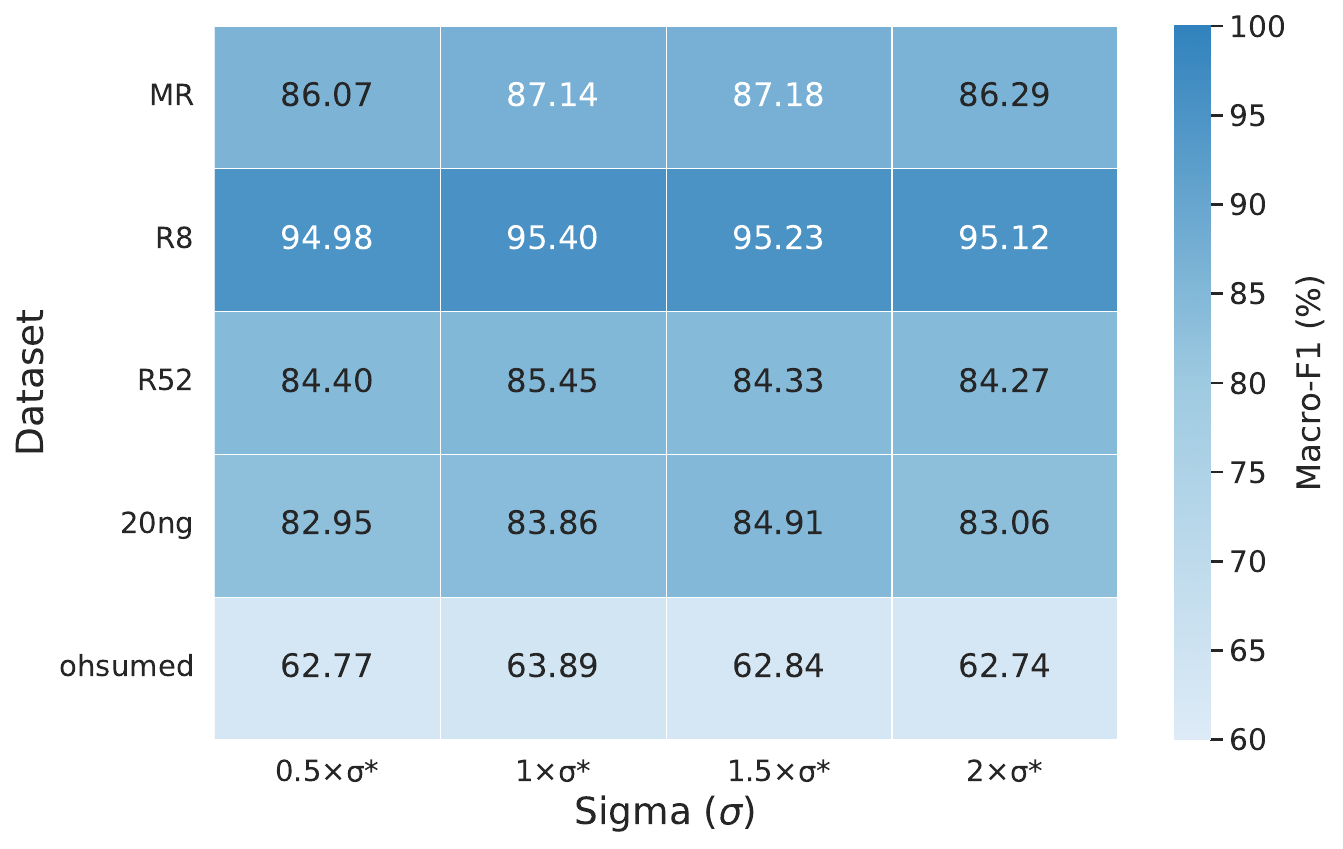}
       \caption{}
       % Sigma parameter $\sigma$ vs Macro-F1
       \label{fig:sigma}
   \end{subfigure}
   \caption{Sensitivity of \textit{G-Loss} (on \texttt{BERT-base-uncased}) to variations in (a) the label-hiding ratio $\gamma$ and (b) the Gaussian kernel width $\sigma$.}
   \label{fig:hyperparameter_analysis}
\end{figure}

\paragraph{\ul{\textbf{Effect of label hiding ratio gamma ($\gamma$) on \textit{G-Loss} performance}}} 
The label hiding ratio, denoted by gamma $(\gamma) \in [0,1]$, plays an important role in \textit{G-Loss} function, by controlling the proportion of labels that remain visible during the LPA iterations. Specifically, a fraction of $(1-\gamma)$ of the labels is masked and LPA is used to infer these hidden labels.    
Figure~\ref{fig:gamma} presents an ablation study on $\gamma $, revealing that intermediate values - particularly in the range $\gamma \sim$ \{$0.5-0.7$\} - consistently lead to better performance across all the datasets. This suggests that revealing approximately $50-70\%$ of the labels encourages the model to generalize effectively by maintaining a balance: enough labeled nodes for stable propagation and enough unlabeled nodes to guide meaningful learning.

\paragraph{\ul{\textbf{Impact of sigma in Gaussian similarity on performance of \textit{G-Loss}}}}
The parameter $\sigma$ in the Gaussian similarity function controls graph connectivity and affects \textit{G-Loss} performance. Small $ \sigma$ value induce sparse, localized neighborhoods that preserve fine-grained class distinctions, while larger values generate denser graphs that may blur class boundaries. 
Figure \ref{fig:sigma} shows performance across $\sigma$ multipliers relative to \texttt{Optuna}-optimized value ($\sigma^*$). We observe dataset-dependent sensitivity: MR and R8 datasets demonstrate remarkable stability, with performance varying by less than $1\%$ across all $\sigma$
values. In contrast, R52 shows a clear optimum at $\sigma^*$ ($85.45\%$), with notable degradation at both extremes ($84.40\%$ at $0.5 \times$ and $84.27\%$ at $2\times$). The 20NG and Ohsumed datasets exhibit intermediate sensitivity, with 20NG peaking at $1.5 \times \sigma^*$ ($84.91\%$) rather than the optimal point.
This analysis reveals two key insights: (1) G-Loss maintains robust performance within a reasonable $\sigma$ range for most datasets and (2) the optimal $\sigma$ balances local precision with global connectivity, underscoring the importance of proper hyperparameter tuning to maximize G-Loss effectiveness.

Additionally, figure~\ref{fig:lambda} presents an ablation study on the weighting factor $\lambda$ in \textit{G-Loss-O}.
% \textit{G-Loss} is most sensitive to $\sigma$, while $\gamma$ and $\lambda$ induce comparatively smaller shifts, indicating stable performance across most configurations.
% \paragraph{\ul{\textbf{Hyperparameter sensitivity analysis of \textit{G-Loss}}}} We further evaluate the robustness of \textit{G-Loss} through a targeted hyperparameter sensitivity analysis on the MR dataset using \texttt{BERT-base-uncased} (Figure~\ref{fig:tornedo}). Taking the optimal configuration as a reference point (Macro-F1 = 0.8714), we perturb each hyperparameter within a small neighborhood while holding the others fixed. The results show that the Gaussian-kernel width ($\sigma$) has the largest influence on performance, yielding a macro-F1 swing of $\pm 0.0111$. In comparison, the loss-weighting coefficient ($\lambda$) and label-hiding ratio ($\gamma$) produce substantially smaller deviations of $\pm 0.0063$ and $\pm 0.0047$, respectively, indicating limited sensitivity to these parameters. Overall, the analysis shows that \textit{G-Loss} is robust to small variations in the hyperparameter values, with only $\sigma$ exhibiting moderate influence on downstream performance.
We further compliment our analysis with a hyperparameter sensitivity study of \textit{G-Loss} using tornado plot in section \ref{app:sensitivity}. To illustrate the qualitative impact of different objectives, we provide t-SNE visualizations of learned embeddings for MR and R8 datasets in section \ref{app:visualizations} of the appendix. These visual analyses offer additional insight into how \textit{G-Loss} influences representation structure compared to traditional loss functions. Finally, we note that evaluating \textit{G-Loss} on extremely large encoder-based models such as DeBERTa-xxlarge (1.5B parameters) was not feasible due to computational limitations. Training such models requires GPU memory far exceeding our available resources, and thus remained outside the scope of this study.

\section{Conclusion}
\label{conclusion}
We proposed a novel graph-guided loss, \textit{G-Loss}, that shifts language model fine-tuning from the local pairwise optimization to global structure alignment. \textit{G-Loss} utilizes a dynamically constructed semantic graph and semi-supervised LPA to capture global document similarity. 
Experiments across five benchmark datasets and three transformer architectures show \textit{G-Loss's} consistent effectiveness in achieving performance improvements over traditional losses. The dynamic mini-batch graph construction ensures computational efficiency and scalability. \textit{G-Loss} highlights the potential of integrating graphs into language model fine-tuning. Future research directions include: (1) Scaling \textit{G-Loss} to large-language models (e.g., GPT, LLaMa), (2) evaluating on larger, non-textual complex datasets, and (3) extending to multi-label and multi-modal tasks. 

\section*{Acknowledgements}
We thank Saurabh Rahul Bhandari, Apoorva Gulati, and Henil Shalin Panchal for their helpful discussions and valuable feedback.

% We would like to express our sincere gratitude to all members who participated for a short duration and played a valuable role in advancing this work. In particular, we thank Saurabh Rahul Bhandari and Apoorva Gulati for their insightful contributions to the experimental design during the initial phase of this study, and Henil Shalin for his valuable input on the mathematical formulation of the \textit{G-Loss-SQRT} approach. Each individual’s efforts and perspectives have meaningfully enriched the quality and depth of this research. 
% For natbib users:
\bibliographystyle{unsrtnat}
\bibliography{reference}
% For bibLaTeX users:
% \printbibliography

\appendix
\clearpage
\begin{table*}[t]
\centering
\renewcommand{\arraystretch}{1.35}
\setlength{\tabcolsep}{6pt}
\caption*{\large\textbf{Appendix Contents}}
\begin{tabular}{@{}p{0.72\linewidth}r@{}}
\toprule
\textbf{Section} & \textbf{Page} \\
\midrule
% ── A  Algorithm ──────────────────────────────────────
\hyperref[app:algorithm]{\textbf{A\; Algorithm for the G-Loss Based Fine-Tuning}}
  & \pageref{app:algorithm} \\[4pt]
% ── B  Traditional Losses ─────────────────────────────
\hyperref[app:losses]{\textbf{B\; Traditional Loss Functions --- Mathematical Formulations}}
  & \pageref{app:losses} \\
\quad\hyperref[app:losses]{\textit{Supervised-Contrastive Loss}}
  & \pageref{app:losses} \\
\quad\hyperref[app:losses]{\textit{Triplet Loss}}
  & \pageref{app:losses} \\
\quad\hyperref[app:losses]{\textit{Cosine-Similarity Loss}}
  & \pageref{app:losses} \\[4pt]
% ── C  Macro-Silhouette ───────────────────────────────
\hyperref[app:silhouette]{\textbf{C\; Macro-Silhouette Coefficient-Based Evaluation Details}}
  & \pageref{app:silhouette} \\[4pt]
% ── D  Sigma Computation ──────────────────────────────
\hyperref[app:sigma]{\textbf{D\; Sigma Value Computation for G-Loss-SQRT}}
  & \pageref{app:sigma} \\
\quad\hyperref[app:sigma]{\textit{First and Second Partial Derivatives}}
  & \pageref{app:sigma} \\
\quad\hyperref[app:sigma]{\textit{Points of Inflection}}
  & \pageref{app:sigma} \\
\quad\hyperref[app:sigma]{\textit{Median Pairwise Distance as Optimal Choice}}
  & \pageref{app:sigma} \\[4pt]
% ── E  Singularity and Stability ──────────────────────
\hyperref[app:stability]{\textbf{E\; Singularity and Stability in the Closed-form LPA Solution}}
  & \pageref{app:stability} \\[4pt]
% ── F  Visualizations ─────────────────────────────────
\hyperref[app:visualizations]{\textbf{F\; Visualizations}}
  & \pageref{app:visualizations} \\[4pt]
% ── G  Hyperparameter Selection ───────────────────────
\hyperref[app:hyperparam]{\textbf{G\; Hyperparameter Selection}}
  & \pageref{app:hyperparam} \\[4pt]
% ── H  Timing Breakdown ───────────────────────────────
\hyperref[app:timing]{\textbf{H\; Detailed Per-Epoch Timing Breakdown}}
  & \pageref{app:timing} \\[4pt]
% ── I  Statistical Significance ───────────────────────
\hyperref[app:significance]{\textbf{I\; Statistical Significance Test on G-Loss}}
  & \pageref{app:significance} \\[4pt]
% ── J  Sensitivity Analysis ───────────────────────────
\hyperref[app:sensitivity]{\textbf{J\; Hyperparameter Sensitivity Analysis of G-Loss}}
  & \pageref{app:sensitivity} \\[4pt]

\midrule
\multicolumn{2}{@{}l}{\textbf{Figures}} \\[2pt]
\hyperref[fig:adj_graph]{Document Similarity and Demo Graph Evolution During Fine-Tuning}
  & \pageref{fig:adj_graph} \\
\hyperref[fig:tsne_comparison]{t-SNE Visualization of Learned Embeddings (G-Loss vs.\ SCL, R8)}
  & \pageref{fig:tsne_comparison} \\
\hyperref[fig:epoch_breakdown]{Training Time Breakdown per Epoch (MR, BERT-base-uncased)}
  & \pageref{fig:epoch_breakdown} \\
\hyperref[fig:tornedo]{Hyperparameter Sensitivity (Tornado Plot, MR)}
  & \pageref{fig:tornedo} \\[4pt]

\midrule
\multicolumn{2}{@{}l}{\textbf{Tables}} \\[2pt]
\hyperref[tab:gloss_hparams]{List of Hyperparameters for G-Loss and Baseline Losses}
  & \pageref{tab:gloss_hparams} \\
\hyperref[tab:gloss_optuna]{G-Loss Hyperparameter Selection and Optimal Range}
  & \pageref{tab:gloss_optuna} \\
\hyperref[tab:stars]{Statistical Significance --- Paired t-Test Results}
  & \pageref{tab:stars} \\[4pt]

\midrule
\multicolumn{2}{@{}l}{\textbf{Algorithms}} \\[2pt]
\hyperref[algo_finetune_function]{Fine-Tuning Process at Minibatch $k$}
  & \pageref{algo_finetune_function} \\
\bottomrule
\end{tabular}
\end{table*}

\clearpage

\section{Algorithm for the \textit{G-Loss} based fine-tuning}
\label{app:algorithm}
The algorithm below presents the outline of our proposed \textit{G-Loss} based fine-tuning approach for language models in the document classification task, executed for each minibatch $k$. The process begins with embedding extraction from the language model (Step 1), followed by normalization (Step 2). Next step is constructing a similarity matrix using a Gaussian kernel function to capture semantic relationships between document embeddings (Step 4). The similarity matrix is then transformed into a normalized adjacency matrix (Step 5) and subsequently partitioned into two sets: labeled subset and masked subset according to the ratio $\gamma$ (Step 6), where the masked labels are hidden from propagation and retained as targets for the loss.
\begin{algorithm}[H]
\caption{Fine-tuning process at minibatch $k$}
\label{algo_finetune_function}
\begin{algorithmic}[1]
\STATE \textbf{Input:}  
$\mathcal{V}_k$: Document set,  
$Y_k$: Label set,  
$\Phi(.)$: language model (LM),  
\STATE \textbf{Hyperparameters:} $\gamma, \eta, \sigma$  
\STATE \textbf{Output:} Optimized weights of LM, $\Phi(.)$  

\STATE  $X_k \gets \Phi(\mathcal{V}_k)$ \COMMENT{Extract embeddings from LM}
\STATE   $X_k \gets X_k / \|X_k\|_2$ \COMMENT{Normalize embeddings}

\STATE  \textbf{Compute similarity matrix:}
\STATE \hskip\algorithmicindent   $W_{ij} \gets \exp\left(-\frac{\|X_{ki} - X_{kj}\|^2}{2\sigma^2}\right) - \text{diag}(W)$

\STATE \hskip\algorithmicindent   $\tilde{A} \gets D^{-1/2} W D^{-1/2}$ \COMMENT{Normalize W}

\STATE  $Y_{kl}, Y_{ku} \gets \gamma$-split$(Y_k)$ \COMMENT{Partition label set}

\STATE \textbf{Label Propagation} 

\hskip\algorithmicindent Transition matrix $\mathbf{T}, \quad \text{where} \quad \mathbf{T_{ij}}\gets \frac{\tilde{A}_{ij}}{\sum_{m=0}^{B} \tilde{A}_{mj} }$

\STATE \hskip\algorithmicindent   $\hat{Y}_{ku} \gets (I - \mathbf{T}_{uu})^{-1}. \mathbf{T}_{ul}Y_{kl}$

\STATE \textbf{Compute cross-entropy loss:}
\STATE \hskip\algorithmicindent   $\mathcal{L_G} \gets -\frac{1}{B_e} \sum\limits_{j=1}^{B_e} \sum\limits_{c=1}^{C} 
            Y_{ku,j}^c \log \big( \hat{Y}_{ku,j}^c \big)$    

% \STATE \textbf{SGD}($\mathcal{L_G}$) \COMMENT{Backpropagate and update LM}
\RETURN Loss $\mathcal{L_G}$ 
\end{algorithmic}
\end{algorithm}
A column-stochastic transition matrix $T$, is formed from the normalized adjacency that represents the probability of transitioning from node $i$ to node $j$ (step 8). Partitioning $T$ into labeled and unlabeled blocks, the soft labels for the masked documents are obtained in closed form as the harmonic solution of LPA.  
The graph loss is the cross-entropy between the true masked labels and
the propagated soft labels, averaged over the $B_e$ masked nodes
and $C$ classes (line~11). The loss grows when the graph fails to recover a masked label, penalizing embeddings whose neighborhood structure disagrees with the true classes. The returned $\mathcal{L}_{\mathcal{G}}$ is backpropagated to update language model, $\Phi(\cdot)$.
\section{Traditional loss functions-mathematical formulations}
\label{app:losses}
\ul{\textbf{Supervised-contrastive loss}}

Supervised contrastive loss (SCL) \citep{scl} enhances traditional contrastive learning \citep{contrastive} by incorporating class labels to optimize embeddings. 
Given a minibatch of samples ${(x_i, y_i)}$ and their representations $(z_i \in \mathbb{R}^d)$), the loss function minimizes intra-class and maximizes inter-class distances in the $(\mathbb{R}^d)$ embedding space.
Mathematically, supervised contrastive loss can be expressed as:

\begin{equation}
\label{eq:L_SCL}
\mathcal{L}_{SCL}(z_i) =  -\frac{1}{|P(i)|} \sum_{p \in P(i)}  \log \frac{\exp(z_i \cdot z_p / \tau)}
{\sum_{a \in A(i)} \exp(z_i \cdot z_a / \tau)}
\end{equation}

where $z$ represents sample embeddings,  
$P(i)$, positive samples for the anchor $z_i$, $A(i)$,all other samples excluding anchor, and
$\tau$, scalar temperature parameter controlling class separation.

\ul{\textbf{Triplet Loss}} Triplet Loss \citep{triplet} constructs triplets (\textit{anchor, positive, negative}) within each minibatch, encouraging the anchor-positive pair to be closer while pushing the negative further away. 
The loss is defined as: 
\begin{equation*}
        \mathcal{L}_{\text{Triplet}} =  
        \frac{1}{N_{bt}} \sum_{i=1}^{N_{bt}}  
        \max \big( 0, \| \mathbf{z}_i - 
        \mathbf{z}_i^+ \|_2^2 - \| \mathbf{z}_i - \mathbf{z}_i^- \|_2^2 + \alpha \big) 
\end{equation*}
where $N_{bt}$ is the number of triplets formed from datapoints in minibatch, and \( z_i, z^+_i, z^-_i \) are the anchor, positive, and negative sample embeddings, respectively, and \( \alpha \) is a margin parameter enforcing separation.  

\ul{\textbf{Cosine-Similarity Loss}}
Cosine-Similarity loss maximizes cosine similarity between positive pairs and minimizes it between negative pairs within a batch. 
Given a minibatch of size \( B \), the loss is defined as:  

\begin{equation*}
    \mathcal{L}_{\text{Cos-sim}} = \frac{1}{N_{p}}\sum_{i=1}^{N_{p}} \left( \frac{z_1^i \cdot z_2^i}{\|z_1^i\| \cdot \|z_2^i\|} - label^i \right)^2
\end{equation*}

where \({N_{p}}\) is the number of generated pairs in minibatch, \(\frac{z_1^i \cdot z_2^i}{||z_1^i|| \cdot ||z_2^i||} \) is cosine similarity between embeddings \( z_1 \) and \( z_2 \), and \( label \in \{0,1\} \) denotes different-class (0) or same-class (1).
\label{mathematical_formula}

\section{Macro-Silhouette coefficient-based evaluation details}
\label{app:silhouette}

In \textit{Macro-Silhouette coefficient} based evaluation strategy, we monitor the cohesiveness of generated embeddings from the learned model. The goal is for the model to learn to discriminate between the embeddings of same-class/different-class documents. 
 
Specifically, at each fine-tuning epoch, we calculate the macro-Silhouette score using embeddings generated from the models' most recent parameters and the true labels of the validation set.
The macro-Silhouette coefficient \cite{macrosil} for a data point \( x_i \) is given by:
\[
s(x_i) = \frac{b(x_i) - a(x_i)}{\max(a(x_i), b(x_i))}
\]

where \( a(x_i) \) is the average intra-class distance, while \( b(x_i) \) is the average nearest-class distance. Silhouette-coefficient for each class, $C_i$ is computed as follow:
\[
S_C = \frac{1}{|C|}\sum_{x_i \in C_i} s(x_i)
\]
And the overall macro-Silhouette coefficient is:

\[ S_{macro} = \frac{1}{K} \sum_{i=1}^{K} S_C(C_i) \]

where \( K \) is the number of classes.

Traditionally, Silhouette score is aggregated using micro-averaging (averaging over all data points), however it can be highly sensitive to cluster imbalance and noise. The Macro-Silhouette score provides greater robustness and reflects the quality of the embedding and the language model's learning progress. We implement early stopping when this score plateaus to capture optimal embedding representations. 

\section{ Sigma value computation for \textit{Gloss-SQRT}}
\label{app:sigma}
The \textbf{Gaussian kernel} is defined as:
\[
k(\mathbf{x}_i, \mathbf{x}_j) = \exp\left(-\frac{\|\mathbf{x}_i - \mathbf{x}_j\|^2}{2\sigma^2}\right).
\]

Let 
\[
d = \|\mathbf{x}_i - \mathbf{x}_j\|^2
\]
(a fixed constant for given $\mathbf{x}_i, \mathbf{x}_j$), so the kernel simplifies to:
\[
k(\sigma) = \exp\left(-\frac{d}{2\sigma^2}\right).
\]

We compute partial derivatives with respect to $\sigma$ (assuming $\sigma > 0$).

\subsection*{1. First Partial Derivative $\frac{\partial k}{\partial \sigma}$}

Rewrite
\[
k = \exp\left(-\frac{d}{2}\sigma^{-2}\right).
\]
Using the chain rule:
\[
\frac{\partial k}{\partial \sigma} = \exp\left(-\frac{d}{2\sigma^2}\right) \cdot \frac{\partial}{\partial \sigma}\left(-\frac{d}{2}\sigma^{-2}\right).
\]

Compute the derivative inside:
\[
\frac{\partial}{\partial \sigma}\left(-\frac{d}{2}\sigma^{-2}\right) = -\frac{d}{2} \cdot (-2)\sigma^{-3} = \frac{d}{\sigma^3}.
\]

Thus:
\[
\boxed{\frac{\partial k}{\partial \sigma} = \frac{d}{\sigma^3} \exp\left(-\frac{d}{2\sigma^2}\right)}.
\]

\subsection*{2. Second Partial Derivative $\frac{\partial^2 k}{\partial \sigma^2}$}

Differentiate $\frac{\partial k}{\partial \sigma}$ using the product rule:
\[
\frac{\partial}{\partial \sigma}\left( \frac{d}{\sigma^3} \exp\left(-\frac{d}{2\sigma^2}\right) \right) 
= d \cdot \frac{\partial}{\partial \sigma}\left( \sigma^{-3} \exp\left(-\frac{d}{2}\sigma^{-2}\right) \right).
\]

Set $u = \sigma^{-3}$ and $v = \exp\left(-\frac{d}{2}\sigma^{-2}\right)$. Then:
\[
\frac{\partial u}{\partial \sigma} = -3\sigma^{-4}, 
\quad \frac{\partial v}{\partial \sigma} = \exp\left(-\frac{d}{2\sigma^2}\right) \cdot \frac{d}{\sigma^3}.
\]

Apply the product rule:
\[
\frac{\partial^2 k}{\partial \sigma^2} = d \left[ \sigma^{-3} \cdot \left( \frac{d}{\sigma^3} \exp\left(-\frac{d}{2\sigma^2}\right) \right) + \exp\left(-\frac{d}{2\sigma^2}\right) \cdot (-3\sigma^{-4}) \right].
\]

Simplify:
\[
\frac{\partial^2 k}{\partial \sigma^2} = d \exp\left(-\frac{d}{2\sigma^2}\right) \left[ \frac{d}{\sigma^6} - \frac{3}{\sigma^4} \right].
\]

\[
\boxed{\frac{\partial^2 k}{\partial \sigma^2} = \frac{d(d - 3\sigma^2)}{\sigma^6} \exp\left(-\frac{d}{2\sigma^2}\right)}.
\]

\subsection*{3. Points of Inflection}

Points of inflection occur where $\frac{\partial^2 k}{\partial \sigma^2} = 0$ or is undefined, and the \textbf{concavity changes}.

\subsubsection*{Critical Points}
\[
\frac{\partial^2 k}{\partial \sigma^2} = 0 \implies d(d - 3\sigma^2) = 0 
\]
(since $\exp(\cdot) > 0$ and $\sigma^6 > 0$ for $\sigma > 0$).

Given $d = \|\mathbf{x}_i - \mathbf{x}_j\|^2 > 0$ (assuming $\mathbf{x}_i \neq \mathbf{x}_j$), solve:
\[
d - 3\sigma^2 = 0 \implies \sigma^2 = \frac{d}{3} \implies \sigma = \sqrt{\frac{d}{3}} \quad (\text{valid since } \sigma > 0).
\]

\subsubsection*{Concavity Change}
- For $\sigma < \sqrt{d/3}$: $d - 3\sigma^2 > 0 \Rightarrow \frac{\partial^2 k}{\partial \sigma^2} > 0$ (concave up).  
- For $\sigma > \sqrt{d/3}$: $d - 3\sigma^2 < 0 \Rightarrow \frac{\partial^2 k}{\partial \sigma^2} < 0$ (concave down).  
- At $\sigma = \sqrt{d/3}$, concavity changes from up to down.  

\textbf{Undefined Point}: $\sigma = 0$ is not in the domain (kernel undefined).

\subsubsection*{Conclusion}
\[
\boxed{\text{Point of inflection at } \sigma = \sqrt{\tfrac{d}{3}} \quad \text{where } d = \|\mathbf{x}_i - \mathbf{x}_j\|^2}
\]

\subsection*{Median Pairwise Distance as Optimal Choice}
The median pairwise distance is often chosen as the kernel bandwidth $\sigma$ because:
\begin{itemize}
    \item It is \textbf{robust to outliers}.
    \item Reflects the \textbf{dominant scale} of the data.
    \item Maximizes \textbf{kernel sensitivity} for typical pairs.
\end{itemize}

\section{Singularity and Stability in the Closed-form LPA Solution}
\label{app:stability}
\textcolor{black}{
The closed-form solution employed in this work follows the classical formulation of graph-based semi-supervised learning~\cite{zhu2003semi,LPA}. When pairwise similarities are computed using the Gaussian kernel, the resulting similarity matrix \( W \) is positive semi-definite. This property ensures that the corresponding normalized transition matrix }
\[
\textcolor{black}{T = D^{-1} W}
\]
\textcolor{black}{is row-stochastic or, in the case of unlabeled data, \emph{substochastic}. The submatrix \( T_{uu} \), representing transitions among unlabeled nodes, therefore satisfies the substochastic property.}

\textcolor{black}{Since \( T_{uu} \) is substochastic, its spectral radius \( \rho(T_{uu}) \) is strictly less than one, i.e.,}
\[
\textcolor{black}{\rho(T_{uu}) < 1.}
\]
\textcolor{black}{This directly implies that the matrix \( (I - T_{uu}) \) is \textbf{non-singular}, ensuring the existence and stability of the closed-form solution.}

\textcolor{black}{Formally, the spectral radius of a square matrix \( A \) is defined as}
\[
\textcolor{black}{\rho(A) = \max_i |\lambda_i|,}
\]
\textcolor{black}{where \( \lambda_i \) denotes the eigenvalues of \( A \). A matrix \( (I - A) \) is invertible if and only if \( 1 \) is not an eigenvalue of \( A \). Consequently, if all eigenvalues of \( A \) satisfy \( |\lambda_i| < 1 \), the matrix \( (I - A) \) is guaranteed to be invertible (i.e., non-singular).}

\textcolor{black}{Therefore, the closed-form label propagation solution,}
\[
\textcolor{black}{F = (I - T_{uu})^{-1} T_{ul} Y_l,}
\]
\textcolor{black}{is well-defined and numerically stable under the practical assumptions governing the mini-batch graph construction used in GLOSS.}

\textcolor{black}{Furthermore, the condition \( \rho(T_{uu}) < 1 \) also guarantees \emph{convergence} of iterative propagation schemes. When this condition holds, the matrix inverse can be equivalently expressed via the Neumann series expansion:}
\[
\textcolor{black}{(I - T_{uu})^{-1} = I + T_{uu} + T_{uu}^2 + T_{uu}^3 + \cdots}
\]
\textcolor{black}{This infinite series converges exactly when the spectral radius of \( T_{uu} \) is less than one. This property provides the theoretical foundation for the convergence of iterative graph-based methods, such as Label Propagation (LPA) and PageRank, as both involve repeated multiplication by a transition matrix whose spectral radius is less than unity.}

\section{Visualizations}
\label{app:visualizations}
\begin{figure*}[htp]
    \centering
    \begin{subfigure}{\textwidth}
        \centering
        \includegraphics[width=\textwidth]{ 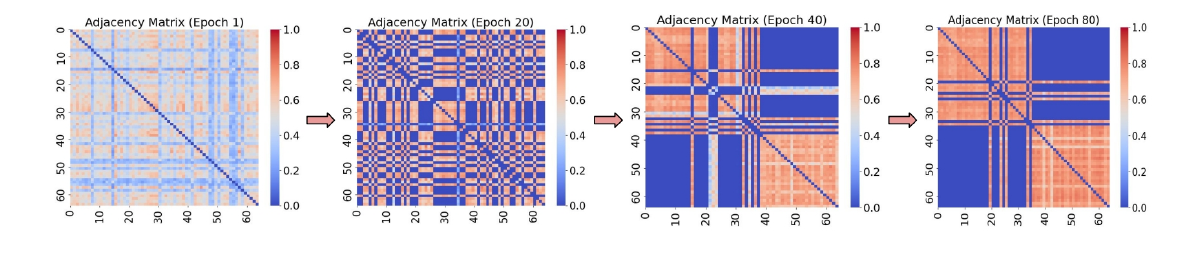}
        \caption{Document similarity evolution for \texttt{Bert-base-uncased} with \textit{G-Loss} fine-tuning. Heatmaps show single minibatch similarity matrices over the epochs.}
        \label{fig:adj_heatmap}
    \end{subfigure}
    
    \begin{subfigure}{\textwidth}
        \centering
        \includegraphics[width=4.8in, height=1.2in]{ 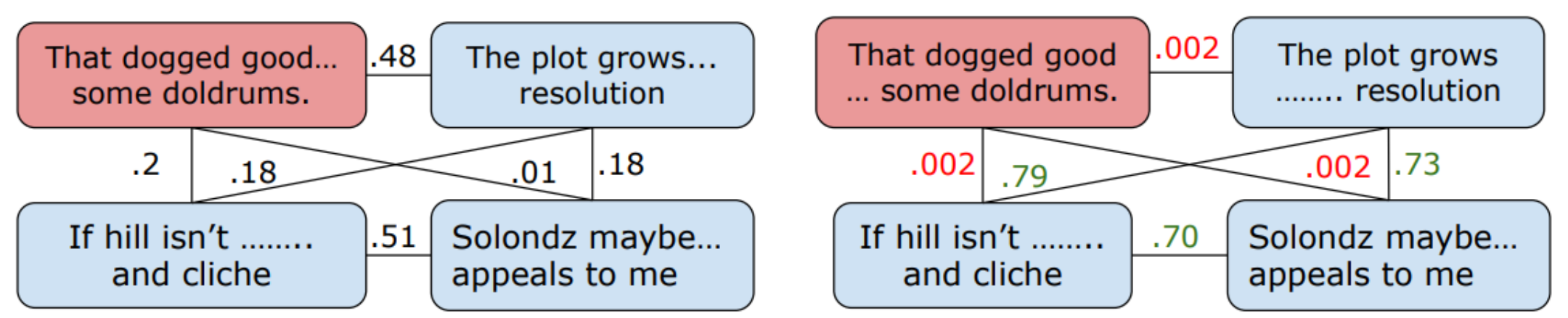}
        \caption{Demo graph evolution during training: the left graph depicts the initial connectivity, while the right graph highlights the strengthened intra-class (green) and weakened inter-class (red) connections, reflecting enhanced semantic coherence and cluster formation.}
        \label{fig:adj_graph}
    \end{subfigure}
\end{figure*}

\textcolor{black}{We visualize the evolution of the document graph during fine-tuning with \textit{G-Loss} using a sample subset from the MR dataset to illustrate how the model progressively refines the underlying semantic structure. Figure~\ref{fig:adj_heatmap} shows the temporal evolution of adjacency heatmaps, capturing how the coherence among same-class data points increases across epochs. Figure~\ref{fig:adj_graph} provides a schematic example of this process: the initial graph (left) appears noisy and poorly aligned with class boundaries, indicating that pre-trained embeddings lack task-specific separation. After fine-tuning (right), the graph exhibits stronger intra-class (green) and weaker inter-class (red) connections. This structural transformation highlights how \textit{G-Loss} promotes the emergence of coherent, class-consistent, and structure-aware representations aligned with the ground-truth labels.}

\begin{figure*}[t]
    \centering
    \begin{subfigure}[b]{0.44\textwidth}
        \centering
        \includegraphics[width=\textwidth]{ 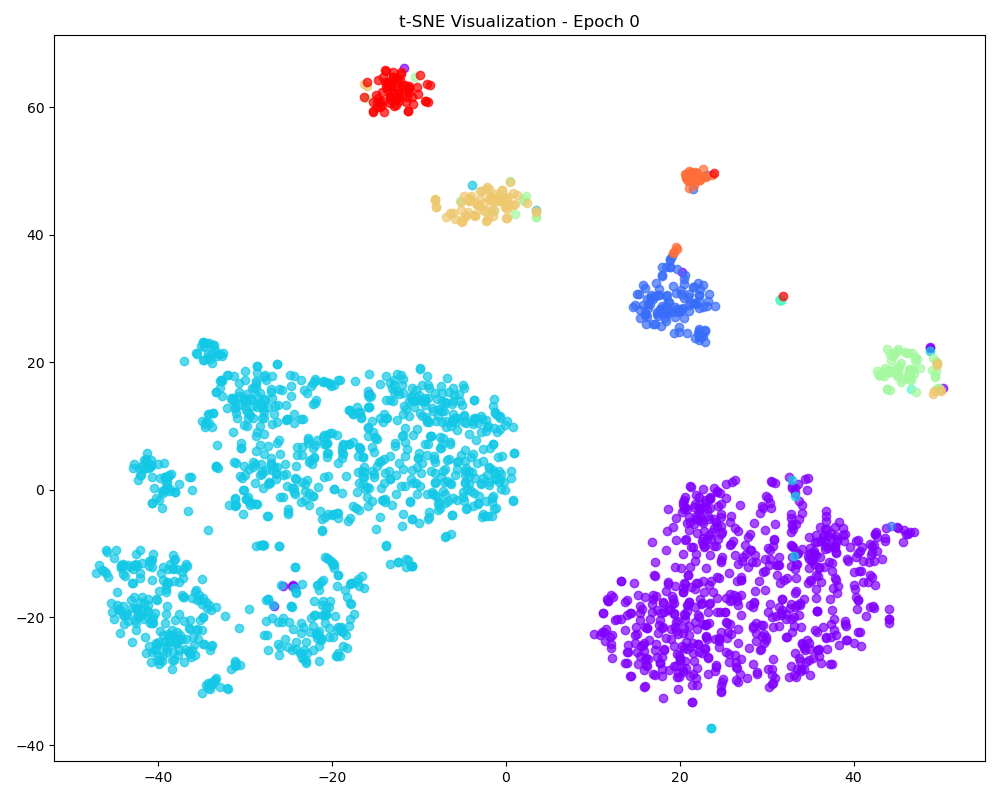}
        \caption{\textit{G-Loss}}
        \label{fig:sub1}
    \end{subfigure}
    \hfill
    \begin{subfigure}[b]{0.44\textwidth}
        \centering
        \includegraphics[width=\textwidth]{ 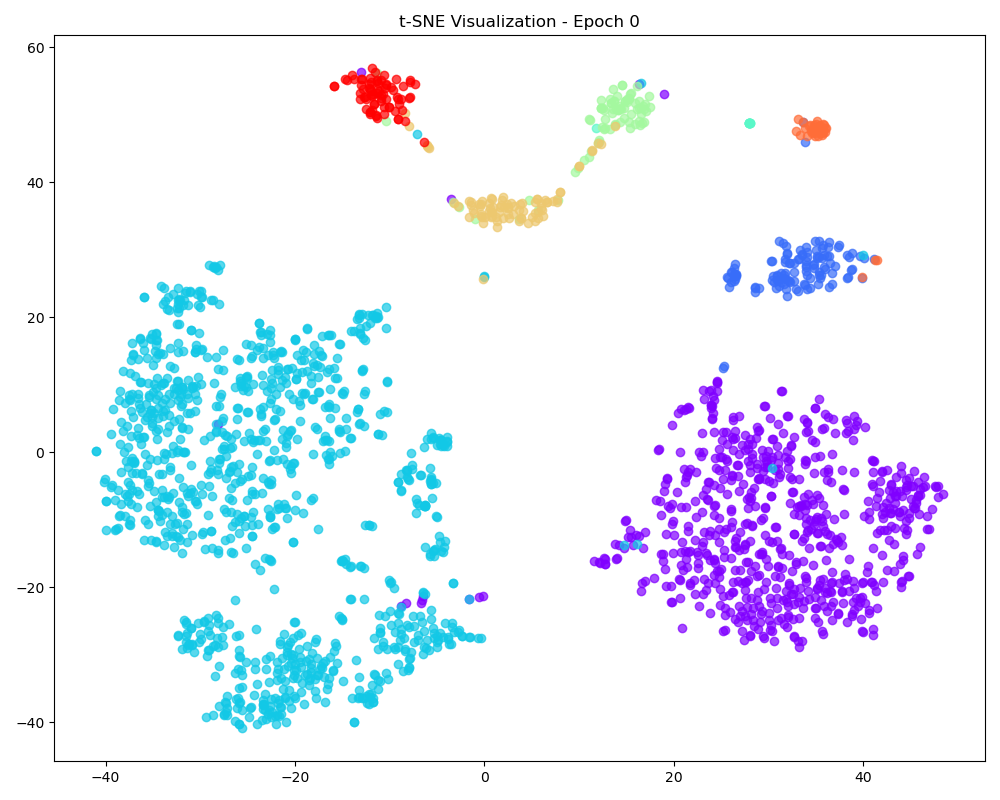}
        \caption{SCL}
        \label{fig:sub2}
    \end{subfigure}
    \caption{\textcolor{black}{t-SNE visualization of learned embeddings with \textit{G-Loss} and SCL on R8 dataset with \texttt{BERT-base-uncased}. \textit{G-Loss} shows a clear separation as compared to SCL.}}
    \label{fig:tsne_comparison}
\end{figure*}

Figure \ref{fig:tsne_comparison} presents t-SNE visualizations of the learned embeddings from \texttt{BERT-base-uncased} model, demonstrating the effectiveness of \textit{G-Loss} in creating well-separated cluster representations. The visualizations reveal distinct class boundaries and improved intra-class cohesion compared to baseline methods. 

\section{Hyperparameter selection}
\label{app:hyperparam}

\textcolor{black}{Table~\ref{tab:gloss_hparams} and Table~\ref{tab:gloss_optuna} summarize the search ranges and empirically optimal intervals obtained via Optuna-based tuning and ablation studies across benchmark datasets. The \textit{G-Loss} framework is controlled by four key hyperparameters: learning rate $(\eta)$, label-hiding ratio $(\gamma)$, weighting coefficient $(\lambda)$, and Gaussian kernel bandwidth $(\sigma)$. Ablation studies show that $\gamma \in [0.5, 0.7]$—hiding about 30-50\% of labels during label propagation—yields the best performance. For the weighting coefficient $\lambda$, experimental results demonstrate that prioritizing the structural regularization term with $\lambda \in [0.7, 0.9]$ enhances geometric alignment in the embedding space. The Gaussian kernel bandwidth $\sigma$ is dataset-dependent and lacks a universal optimal range, necessitating data-specific tuning. To address this challenge, we introduce two \textit{G-Loss} variants: \textit{G-Loss-SQRT} employs an analytical approximation for $\sigma$ derived from the methodology presented in Section~\ref{app:sigma}, offering computational efficiency for resource-constrained deployments; conversely, \textit{G-Loss-O} uses \texttt{Optuna} to identify the optimal $\sigma$, for maximal performance at the cost of increased hyperparameter tuning overhead. This dual design allows practitioners to balance computational cost and accuracy as needed.}

\textcolor{black}{Additionally, Figure~\ref{fig:lambda} presents an ablation study on the weighting factor $\lambda$, which effectively modulates the trade-off between local label supervision via the CE loss and global alignment of the embedding structure using \textit{G-Loss}.Our results show that a range of ${0.7-0.9}$ consistently yields strong performance across benchmark datasets, indicating that this weighting optimally integrates supervised signals and graph-structured relational information, thereby enhancing both predictive accuracy and embedding coherence.}
\begin{table}[H]
\centering
\begin{minipage}{0.45\textwidth}
\centering
\caption{\textcolor{black}{List of hyperparameters for \textit{G-Loss} and other baseline losses.}}
\begin{tabular}{@{}ll@{}}
\toprule
\textbf{Loss} & \textbf{Hyperparameters} \\ \hline
Cross Entropy & Learning rate ($\eta$) \\ \hline
Cosine-Similarity & Learning rate ($\eta$) \\ \hline
Triplet & Learning rate ($\eta$) \\ \hline
CE + SCL & \begin{tabular}[l]{@{}l@{}}Learning rate ($\eta$), \\  Temperature ($\tau$), \\ Weight factor($\lambda$)\end{tabular} \\ \hline
CE + GLoss-SQRT & \begin{tabular}[l]{@{}l@{}}Learning rate ($\eta$), \\  Label-hiding ratio ($\gamma$), \\ Weight-factor ($\lambda$)\end{tabular} \\ \hline
CE + GLoss-O & \begin{tabular}[l]{@{}l@{}}Learning rate ($\eta$), \\ Label-hiding ratio ($\gamma$), \\ Weight-factor ($\lambda$),\\ Gaussian-kernel width ($\sigma$)\end{tabular} \\ \bottomrule
\end{tabular}

\label{tab:gloss_hparams}
\end{minipage}
\hfill
\begin{minipage}{0.5\textwidth}
\centering
\caption{\textcolor{black}{G-Loss Hyperparameter Selection and Optimal Range: For the hyperparameter $\sigma$, we propose a resource-constrained approach, CE+G-Loss-SQRT, where the value of $\sigma$ can be determined mathematically, eliminating the need for Optuna-based tuning.}}
\begin{tabular}{lll}
\hline
\textbf{} & \textbf{\begin{tabular}{l} Optuna\\search range \end{tabular}} & \textbf{\textbf{\begin{tabular}{l} Optimal\\ range \end{tabular}}}  \\
\hline
($\eta$) & \textbf{\begin{tabular}{l} [$ 1e-05, 2e-05$,\\ $3e-05,4e-05,$\\$5e-05]$ \end{tabular}} &  - \\
\hline
($\gamma$) & $0.1-0.9$ &$ 0.5-0.7$ \\
\hline
($\sigma$) & $0.01-10.0$ & Data-specific  \\
\hline
($\lambda$) & $0.1-0.9 $& $0.7-0.9$  \\
\hline
\end{tabular}

\label{tab:gloss_optuna}
\end{minipage}
\end{table}

\section{Detailed per epoch timing breakdown}
\label{app:timing}
\begin{figure}[]
    \centering
    \includegraphics[width=0.99\linewidth]{ 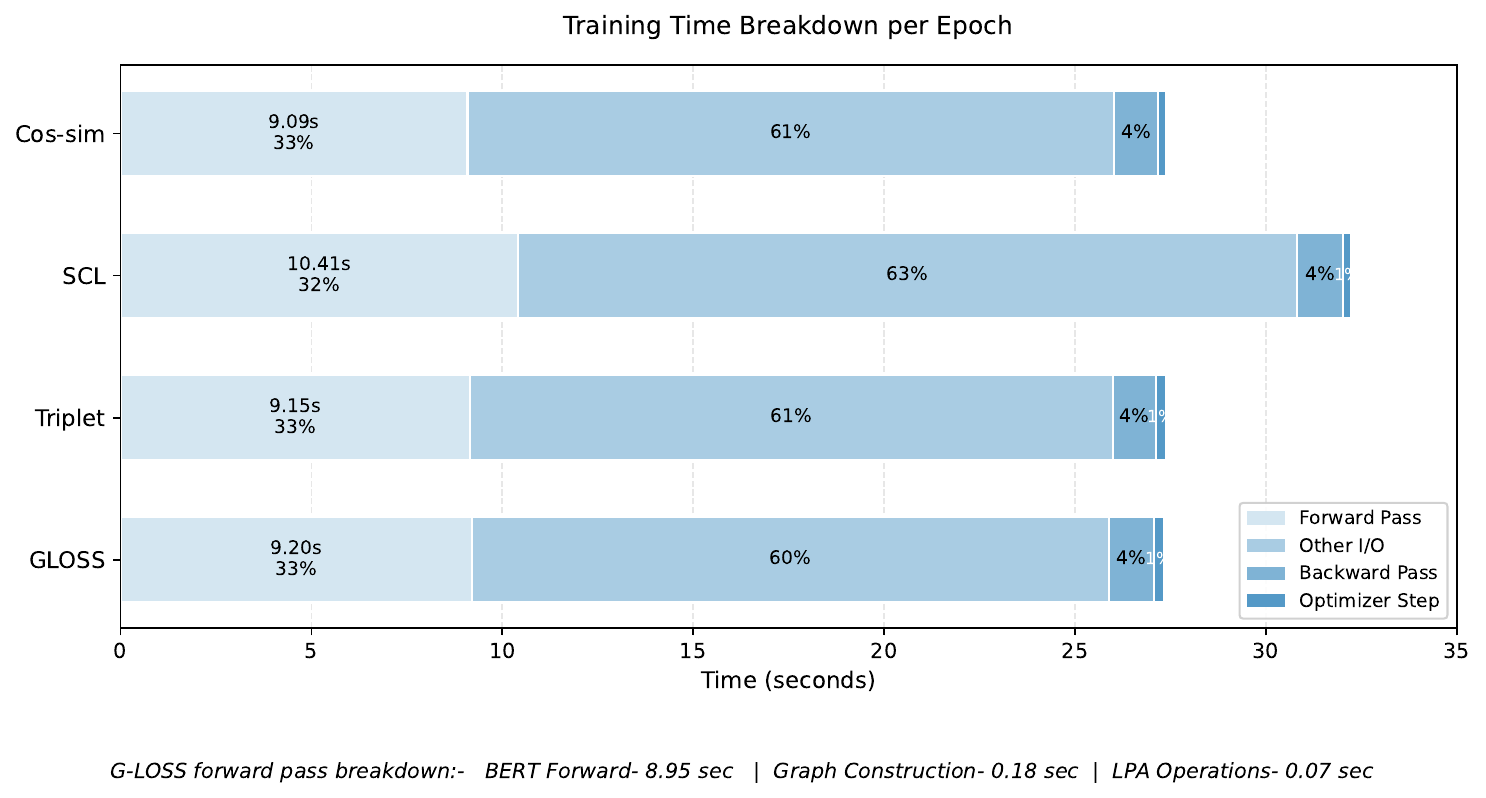}
    \caption{\textcolor{black}{Training time breakdown per epoch across different loss functions on the MR dataset with \texttt{BERT-base-uncased} model. Each horizontal bar represents the total epoch time (in seconds) decomposed into four components: forward pass, other I/O operations, backward pass, and optimizer step. The percentages indicate each component's contribution to the total epoch time. Despite incorporating graph construction and label propagation operations, \textit{G-Loss} maintains comparable computational efficiency to baseline methods.}}
    \label{fig:epoch_breakdown}
\end{figure}

\textcolor{black}{The figure~\ref{fig:epoch_breakdown} presents a detailed per-epoch timing breakdown for different training configurations - \textit{G-Loss}, Triplet, SCL, and Cos-sim — highlighting the relative computational cost across major components: forward pass, backward pass, optimizer step, and I/O operations.
\textit{G-Loss} requires $27.74$ seconds per epoch, nearly identical to Triplet loss ($27.58$ sec) and Cosine similarity ($27.76$ sec), while being $15\%$ faster than SCL ($32.62$ sec). The forward pass accounts for around $32-33\%$ of the computation across all loss functions. Notably, \textit{G-Loss's} graph-based components add minimal overhead: graph construction consumes only $0.18$ seconds ($0.7\%$ of total time) and LPA operations require $0.07$ seconds ($0.2\%$ of total time). These results demonstrate that integrating structural loss introduces minimal computational burden while maintaining efficiency comparable to that of conventional training pipelines.}

% \begin{figure}[]
%     \centering
%     \includegraphics[width=\linewidth]{ heatmap_lambda.pdf}
%     \caption{\textcolor{black}{Hyperparameter sensitivity analysis of G-Loss on \texttt{BERT-base-uncased}: loss weighting factor Lambda $(\lambda)$ vs performance}}
%     \label{fig:lambda}
% \end{figure}

\begin{table}[t]
\centering
\caption{Mean accuracies of \textit{G-Loss} across benchmark datasets compared with alternative losses (SCL, Cosine Similarity, and Triplet). The table reports absolute performance differences and paired t-test p-values, demonstrating that \textit{G-Loss} offers statistically significant improvements $(p < 0.05)$ across all comparisons, highlighting its robustness and consistent advantage.
Significance stars represent: ** - $p<0.05$, *** - $p<0.01$, **** - $p<0.001$.
}
\begin{tabular}{@{}lccccc@{}}
\toprule
Comparison & $\mu$(G-Loss) & $\mu$(Baseline) & $\Delta\mu$ & $p$-value & Observations \\
\midrule
G-Loss vs SCL     & 87.05 & 86.41 & 0.63 & 0.006239 & *** \\
G-Loss vs Cos-Sim & 87.05 & 86.32 & 0.73 & 0.000493 & **** \\
G-Loss vs Triplet & 87.05 & 86.50 & 0.55 & 0.019647 &  ** \\
\bottomrule
\end{tabular}
\label{tab:stars}
\end{table}

\section{Statistical significance test on \textit{G-Loss}}
\label{app:significance}
Table \ref{tab:stars} demonstrates that \textit{G-Loss} outperforms baselines across all benchmark datasets. Relative to SCL, Cosine Similarity, and Triplet objectives, \textit{G-Loss} yields absolute accuracy gains of $0.63$, $0.73$, and $0.55$ points, respectively. The statistical significance of these improvements, as indicated by paired t-tests ($p < 0.05$ in each case), underscores the robustness of \textit{G-Loss} in leveraging graph-based label propagation for enhanced classification performance. The significance-star patterns (marked with *** for $p < 0.01$ and **** for $p < 0.001$) confirm that the performance gains are not merely due to random variation. This consistent performance advantage, coupled with strong statistical significance, validates \textit{G-Loss} as an effective training objective for fine-tuning language models.

\begin{figure}
    \centering
    \includegraphics[width=0.7\linewidth]{ 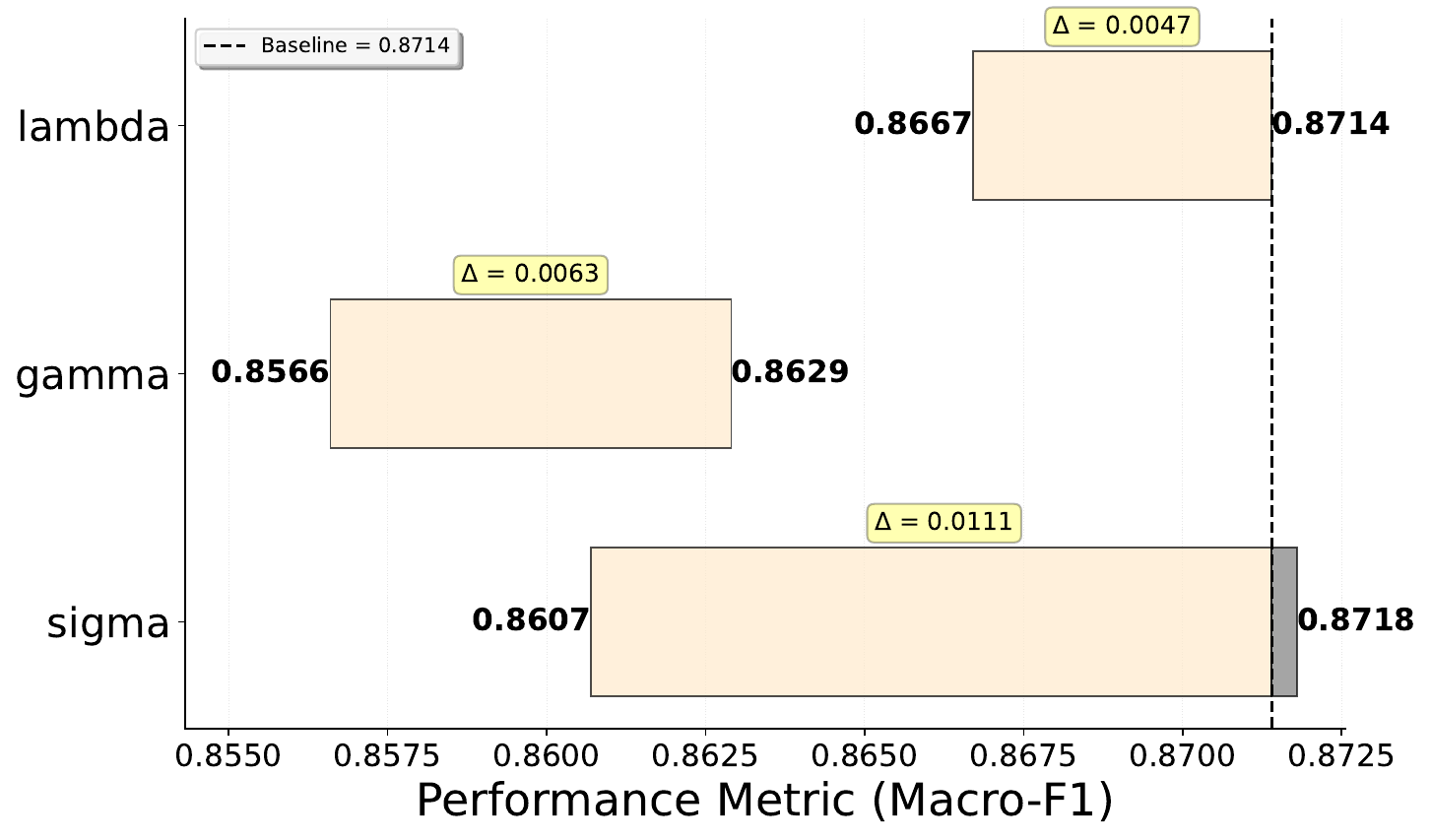}
    \caption{Hyperparameter sensitivity analysis of \textit{G-Loss} on MR dataset with \texttt{BERT-base-uncased}. Each bar represents the maximum macro-F1 deviation when perturbing a single hyperparameter around its optimal value.}
    \label{fig:tornedo}
\end{figure}

% \begin{figure*}[t]
%     \centering
%     \begin{subfigure}[b]{0.48\textwidth}
%         \includegraphics[width=\linewidth]{ heatmap_lambda.pdf}
%     \caption{}
%     \label{fig:lambda}
%     \end{subfigure}
%     \hfill
%     \begin{subfigure}[b]{0.48\textwidth}
%         \centering
%         \includegraphics[width=\linewidth]{ tornado_mr.pdf}
%     \caption{}
%     \label{fig:tornedo}
%     \end{subfigure}
% \caption{(a) Loss weighting factor Lambda $(\lambda)$ vs performance of \textit{G-Loss} on \texttt{BERT-base-uncased} (b) Hyperparameter sensitivity analysis of \textit{G-Loss} on MR dataset with \texttt{BERT-base-uncased}. Each bar represents the maximum macro-F1 deviation when perturbing a single hyperparameter around its optimal value.}
% \end{figure*}

\section{Hyperparameter sensitivity analysis of \textit{G-Loss}}
\label{app:sensitivity}
We further evaluate the robustness of \textit{G-Loss} through a targeted hyperparameter sensitivity analysis on the MR dataset using \texttt{BERT-base-uncased} (Figure~\ref{fig:tornedo}). Taking the optimal configuration as a reference point (Macro-F1 = 0.8714), we perturb each hyperparameter within a small neighborhood while holding the others fixed. The Gaussian kernel width ($\sigma$) emerges as the dominant factor, inducing a performance variation of $\pm 0.0111$. In comparison, the loss-weighting coefficient ($\lambda$) and label-hiding ratio ($\gamma$) produce substantially smaller deviations of $\pm 0.0063$ and $\pm 0.0047$, respectively, indicating limited sensitivity to these parameters. Overall, the analysis shows that \textit{G-Loss} is robust to small variations in the hyperparameter values, with only $\sigma$ exhibiting moderate influence on downstream performance.

\end{document}